\documentclass{article} % For LaTeX2e
\usepackage{collas2023_conference,times}
\usepackage{easyReview}

\usepackage{microtype}
\usepackage{graphicx}
\usepackage{subfigure}
\usepackage{booktabs} % for professional tables
\usepackage{natbib}
\usepackage{caption}
\usepackage{mathtools}

\usepackage[utf8]{inputenc} % allow utf-8 input
\usepackage[T1]{fontenc}    % use 8-bit T1 fonts
\usepackage{url}            % simple URL typesetting
\usepackage{nicefrac}       % compact symbols for 1/2, etc.

\usepackage{comment}
\usepackage{xcolor}
\usepackage{bbm}
\usepackage{verbatim}
\usepackage[linesnumbered, ruled,vlined, algo2e]{algorithm2e}
\usepackage{algorithm,algorithmic}
\usepackage{enumitem}
\usepackage{breqn} % https://tex.stackexchange.com/questions/21290/how-to-make-left-right-pairs-of-delimiter-work-over-multiple-lines

\usepackage{pifont} % check mark \ding{52}

\usepackage{amssymb}
\let\emptyset\varnothing

\setlist[itemize]{noitemsep, topsep=0pt}
\usepackage{tikz}
\usetikzlibrary{intersections,shapes.arrows}

\definecolor{richardcolor}{rgb}{0.2, 0.4, 0.0}

\usepackage{amsthm,amsmath,amssymb, amsfonts}

\usepackage{wrapfig}

\usepackage{titletoc}

\allowdisplaybreaks

\theoremstyle{plain}

\theoremstyle{definition}

\theoremstyle{remark}

\usepackage{todonotes}
\newcommand{\piotrm}[1]{\textcolor{blue}{\small [pm: #1]}}

\newcommand{\sam}[1]{\textcolor{red}{\small [sk: #1]}}

\usepackage{amsmath,amsfonts,bm}

\def\eqref#1{equation~\ref{#1}}
\def\1{\bm{1}}

\def\rvx{{\mathbf{x}}}

\def\vx{{\bm{x}}}

\DeclareMathAlphabet{\mathsfit}{\encodingdefault}{\sfdefault}{m}{sl}
\SetMathAlphabet{\mathsfit}{bold}{\encodingdefault}{\sfdefault}{bx}{n}

\newcommand{\normltwo}{L^2}

\usepackage{hyperref}
\hypersetup{
    colorlinks=true,
    linkcolor=red,
    filecolor=magenta,
    urlcolor=blue,
    citecolor=purple,
    pdftitle={Overleaf Example},
    pdfpagemode=FullScreen,
    }

\usepackage[capitalize,noabbrev,nameinlink]{cleveref}
\usepackage[capitalize,nameinlink]{cleveref}

\usepackage{cancel}

\usepackage[title]{appendix}
\crefname{appsec}{appendix}{appendices}
\Crefname{appsec}{Appendix}{Appendices}

\title{The Effectiveness of World Models for \\ 
        Continual Reinforcement Learning}

\author{Samuel Kessler\textsuperscript{1}, Mateusz Ostaszewski\textsuperscript{2}, Micha\l{} Bortkiewicz\textsuperscript{2}, Mateusz \.Zarski,\textsuperscript{3} \\
\textbf{Maciej Wo\l{}czyk\textsuperscript{4}, Jack Parker-Holder\textsuperscript{1}, Stephen J. Roberts\textsuperscript{1}, Piotr Miłoś\textsuperscript{5}}\\
\\
\textsuperscript{1} University of Oxford \\
\textsuperscript{2} Warsaw University of Technology \\
\textsuperscript{3} 
Institute of Theoretical and Applied Informatics, PAS \\
\textsuperscript{4} Jagiellonian University, Cracow
\\
\textsuperscript{5} Polish Academy of Sciences and Ideas NCBR \\
\\
\texttt{skessler@robots.ox.ac.uk}
}

\collasfinalcopy % Uncomment for camera-ready version, but NOT for submission.

\begin{document}

\maketitle

\begin{abstract}
World models power some of the most efficient reinforcement learning algorithms. In this work, we showcase that they can be harnessed for continual learning – a situation when the agent faces changing environments. World models typically employ a replay buffer for training, which can be naturally extended to continual learning. We systematically study how different selective experience replay methods affect performance, forgetting, and transfer. We also provide recommendations regarding various modeling options for using world models. The best set of choices is called Continual-Dreamer, it is task-agnostic and utilizes the world model for continual exploration. Continual-Dreamer is sample efficient and outperforms state-of-the-art task-agnostic continual reinforcement learning methods on Minigrid and Minihack benchmarks.

\end{abstract}

\section{Introduction}

 \begin{wrapfigure}{r}{0.35\textwidth}
    \centering
    \vspace{-1.7cm}
    %\hspace{0.5cm}
    \includegraphics[width=0.3\textwidth]{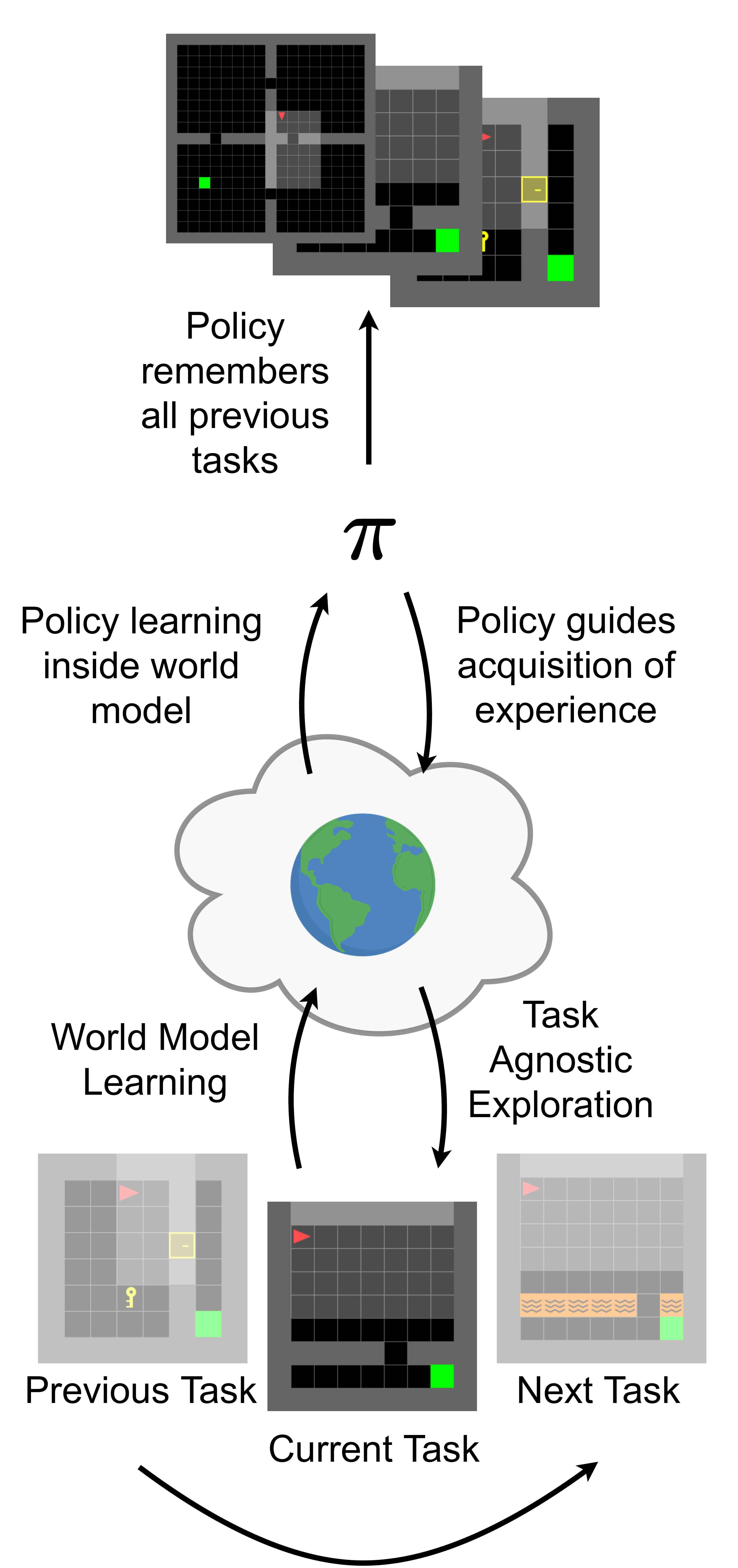}
    \caption*{CRL with world models.}
    \label{fig:mockup}
    \vspace{-1.2cm}
\end{wrapfigure}

\looseness-1 There have been many recent successes in reinforcement learning (RL), such as in games~\citep{muzero}, robotics~\citep{dexterity} and in scientific applications \citep{nguyen2021deep, degrave2022magnetic}. However, these successes showcase methods for solving individual tasks. Looking beyond, the field of continual reinforcement learning (CRL) aims to develop agents which can solve many tasks, one after another, continually while retaining performance on all previously seen ones~\citep{crl_review}. Such capabilities are conjectured to be essential for truly scalable intelligent systems,~\cite{ring1994continual} and~\cite{HASSABIS2017245} conjecture that truly scalable intelligent systems will additionally need to master many tasks in a continual manner. The field of continual reinforcement learning (CRL) aims to develop agents which can solve many tasks, one after another, continually while retaining performance on all previously seen ones~\citep{crl_review}.

\emph{World models} combine generative models with RL and have become one of the most successful paradigms in single-task RL~\citep{ha2018world}. World models are typically trained iteratively, first, the policy interacts with the environment, second, the collected experience is used to train the world model and then the policy is trained using hallucinated experience generated by the world model~\citep{kaiser2019model}. Such approaches are typically sample efficient, offloading the most data-hungry trial-and-error RL training to the imagined world. Further, generative models can be used to create compact representations that facilitate policy training and so obtain very good results for challenging pixel-based environments~\cite {hafner2023mastering}.

This paper explores using world models for learning tasks sequentially. We showcase a method that satisfies the traditional CL desiderata: avoids {catastrophic forgetting}, achieves transfer, high average performance, and is scalable. The proposed method \emph{Continual-Dreamer} is built on top of DreamerV2~\citep{hafner2020mastering} -- an RL algorithm, which achieves state-of-the-art on a number of benchmarks.  We define the Continual-Dreamer configuration in~\cref{sec:experiments}. DreamerV2 uses replay buffer~\citep{experience_replay}, which we extend across many tasks to mitigate forgetting akin to \citep{isele2018selective, clear}. Additionally, we demonstrate that world models are capable to operate without explicit task identification. This is an important requirement in many CRL scenarios and enables further capabilities. In particular, we implement an adaptive exploration method similar to \citep{steinparz2022reactive}, which implicitly adapts to task changes. Based on our empirical results we argue that world-models offer a potent approach for CRL and should attract more research attention.

% , which is another important requirement for CRL.% We also show that world models can be used to enable task-agnostic exploration~\citep{steinparz2022reactive}, which is a key requirement for CRL. See~\cref{fig:mockup} for an overview of our approach. So world models are a good choice for CRL and this work is the first to explore them.

% At the same time, it is \textit{task-agnostic}, namely, it does not require external task identification.
%  Also, we show that world models do not require knowledge of when tasks change and can be \emph{task-agnostic}, the world model's recurrent architecture enables it to be task-agnostic~\citep{caccia2022task}. They also admit a task-agnostic exploration method~\citep{steinparz2022reactive} this is significant since they do not require a new $\epsilon$-greedy exploration schedule or new entropy regularizer when the agent interacts with a new environment. See~\cref{fig:mockup} for an overview of our approach. So world models are a good choice for CRL and this work is the first to explore them.

%\sam{Add references to \emph{Continual-Dreamer} and \emph{Continual-Dreamer} + Plan2Explore throughout the paper.}

Our contributions are as follows\footnote{Code available: \textcolor{purple}{\url{https://github.com/skezle/continual-dreamer}}}:
\begin{itemize}
    \item We present the first approach to task-agnostic model-based CRL. We use DreamerV2 as a backbone, and our work is transferable to other world models.
    \item We evaluate our method performance on two challenging CRL benchmarks, Minigrid and Minihack, and show that it outperforms state-of-the-art task-agnostic methods, demonstrating that the model-based paradigm is a viable solution to CRL.
    \item We thoroughly explore different experience replay strategies, which address how we sample from or populate the experience replay buffer to balance preventing forgetting of previously learned skills while enabling learning new skills from new tasks. 
\end{itemize}

\section{Preliminaries}

\subsection{Reinforcement Learning}
\newcommand{\MDP}{$\mathrm{MDP}$}
\newcommand{\POMDP}{$\mathrm{POMDP}$}

 A Partially Observable Markov Decision Process (\POMDP{}~\citep{kaelbling1998planning}) is the following tuple $(\mathcal{S},\mathcal{A},P,R, \Omega, \mathcal{O}, \gamma)$. Here, $\mathcal{S}$ and $\mathcal{A}$ are the sets of states and actions respectively, such that for $s_t, s_{t+1} \in \mathcal{S}$ and $a_t \in \mathcal{A}$. $P(s_{t+1}| s_t, a_t)$ is the transition distribution and $R(a_t,s_t,s_{t+1})$ is the reward function. Additionally, $\gamma \in (0,1)$ is the discount factor. Since the environments we consider are partially observable, the agent does not have access to the environment state $s \in \mathcal{S}$, but only the observations $o \in \Omega$, where $\Omega$ is the set of observations and $\mathcal{O}: \mathcal{S} \times \mathcal{A} \rightarrow P(\Omega)$ is an observation function that defines a distribution over observations. Actions $a_t$ are chosen using a policy $\pi$ that maps observations to actions: $\Omega \rightarrow\mathcal{A}$. For the purposes of this introduction, let us assume we have access to the states $s_t$ and we are working with a finite horizon $H$. Then the return from a state is $R_t = \sum_{i=t}^H \gamma^{(i \text{-} t)}r(s_i, a_i)$. In RL the objective is to maximize the expected return $J = \mathbb{E}_{a_i \sim \pi, s_0 \sim \rho}[R_1|s_0]$ where $s_0 \sim \rho(s_0)$ and $\rho(\cdot)$ is the initial state distribution.
 %  \maciej{I see what you mean, but it kind of sounds as if the actor has the access to the states}\sam{added \emph{For the purposes of this introduction} so that it is clear that in general we do not have access to $s_t$ for the next sections.}

One approach to maximizing expected return is to use a \emph{model-free} approach, to learn a policy $\pi_\phi:\mathcal{S} \rightarrow \mathcal{A}$ with a parametric model such as a neural network with parameters $\phi$ guided by an action-value function $Q_\theta(s_t, a_t)$ with parameters $\theta$. Alternatively, instead of learning a policy directly from experience we can employ \emph{model-based} RL (MBRL) and learn an intermediate model $f$, for instance, a transition model $s_{t+1} = f(s_t, a_t)$ from experience and learn our policy with additional experience generated from the model $f$~\citep{sutton1991dyna}. Instead of working with the actual state $s_t$, our methods consider the observations $o_t$, we employ recurrent policies, action-value functions, and models to help better estimate states $s_t$~\citep{hausknecht2015deep}.

\subsection{Continual Reinforcement Learning}

In continual RL the agent has a budget of $N$ interactions with each task environment $\mathcal{T}_{\tau}$ The agent is then required to learn a policy to maximize rewards in this environment, before interacting with a new environment and having to learn a new policy. Each task is defined as a new \POMDP{}, $\mathcal{T}_{\tau} = \big( \mathcal{S}_{\tau},\mathcal{A}_{\tau},P_{\tau},R_{\tau}, \Omega_{\tau}, \mathcal{O}_{\tau}, \gamma_{\tau} \big)$. See~\cref{app:cl_def} for a definition of CL in the supervised learning setting.
% \maciej{I guess that depends on the formalism?} \sam{In the next paragraph we explain that the formalize is the setting which is most widely used in the current literature.}. \maciej{Okay, sounds good!}

The agent is continually evaluated on all past and present tasks and so it is desirable for the agent's policy to transfer to new tasks while not forgetting how to perform past tasks. CRL is not a new problem setting~\citep{THRUN199525}, however, its definition has evolved over time and some settings will differ from paper to paper, we employ the setting above which is related to previous recent work in CRL~\citep{ewc, progress_compress, clear,wolczyk2021continual,kessler2021same,powers2021cora, caccia2022task}.

\textbf{Assumption on the task distribution.} In this work we set out to study how world models deal with changing states spaces and transition distributions. We assume that for all pairs of tasks $\mathcal{T}_{i}$ and $\mathcal{T}_{j}$ that the state-spaces between tasks are disjoint: $\forall (\mathcal{T}_{i}, \mathcal{T}_{j} ), \, \mathcal{S}_{i} \cap \mathcal{S}_{j} = \emptyset$.\footnote{Since we are working with \POMDP{}s certain observations from different tasks might coincide.} This setting is popular for evaluating CRL benchmarks~\citep{wolczyk2021continual, powers2021cora}.

\section{Related Work}

% \subsection{Continual Supervised Learning}

% Here, we briefly describe CL methods. One approach to CL referred to as \emph{regularization approaches} regularizes a NN's weights to ensure that optimizing for a new task finds a solution that is ``close'' to the previous task's \citep{ewc, vcl, Zenke2017}. Working with functions can be easier than with NN weights and so task functions can be regularized to ensure that learning new function mappings are ``close'' across tasks \citep{Li2017, Benjamin2019, buzzega2020dark}. By contrast, \emph{expansion approaches} add new NN components to enable learning new tasks while preserving components for specific tasks \citep{progressivenets, lee2020neural}. \emph{Memory approaches} replay data from previous tasks when learning the current task. This can be performed with a generative model \citep{Shin2017}. Or samples from previous tasks (\emph{memories}) \citep{Lopez-Paz, Aljundi2019, ChaudhryTinyEps}. 

\textbf{Continual Reinforcement Learning}. Seminal work in CRL, EWC~\citep{ewc} enables DQN~\citep{Mnih} to continually learn to play different Atari games with limited forgetting. EWC learns new Q-functions by regularizing the parameter-weighted L2 distance between the new task's current weights and the previous task's optimal weights. EWC requires additional supervision informing it of task changes to update its objective, select specific Q-function head and select a task-specific $\epsilon$-greedy exploration schedule. Progress and Compress~\citep{progress_compress} applies a regularization to policy and value function feature extractors for an actor-critic approach. Alternatively, LPG-FTW \citep{Mendez2020} learns an actor-critic that factorizes into task-specific parameters and shared parameters. Both methods require task supervision and make use of task-specific parameters and shared parameters. Task-agnostic methods like CLEAR~\citep{clear} do not require task information to perform CRL. CLEAR leverages experience replay buffers~\citep{experience_replay} to prevent forgetting: by using an actor-critic with V-trace importance sampling~\citep{espeholt2018impala} of past experiences from the replay buffer. Model-based RL approaches to CRL have been demonstrated where the model weights are generated from a hypernetwork which itself is conditioned by a task embedding~\citep{huang2021continual}. Recent work demonstrates that recurrent policies for POMDPs can obtain good overall performance on continuous control CRL benchmarks~\citep{caccia2022task}. Another task-aware solution is to expand a subspace of policies, the number of policies scaling sublinearly with the number of tasks~\citep{gaya2022building}. Of the related works presented, only CLEAR is task-agnostic and so is the primary baseline under consideration when comparing task-agnostic world model approaches to CRL. A number of previous works have studied transfer in multi-task RL settings where the goals within an environment change~\citep{barreto2017successor,schaul2015universal,barreto2019option}. In particular, incorporating the task definition directly into the value function~\citep{schaul2015universal} and combining this with off-policy learning allows a CRL agent to solve multiple tasks continually, and generalize to new goals~\citep{mankowitz2018unicorn}. Model-based RL mathods have also been assessed by looking at how quickly they can adapt to changes in reward function~\citep{van2020loca}. When only the reward function changes between tasks then experience from the replay buffer can interfere to prevent learning a new task or cause forgetting \citep{wan2022towards}. The problem of interference can be mitigated by having separate policy heads per task~\citep{kessler2021same}. See~\cref{app:cl_related_works} for related works on continual supervised learning.

\textbf{Continual Adaptation.} Instead of focusing on remembering how to perform all past tasks, another line of research investigates quick adaptation to changes in the environment. This can be captured by using a latent variable and off-policy RL~\citep{xie2020deep}. Alternatively, one can meta-learn a model such that it can then adapt quickly to new changes in the environment~\citep{nagabandi2018deep}. All these works use small environment changes such as modification of the reward function or variations in gravity or mass of certain agent limbs as new tasks. The tasks which we consider in this work contain substantially different $\mathcal{A}$, $\mathcal{S}$ from one task to the next. For example, skills such as opening doors with keys or avoiding lava, or crossing a river which are quite different in comparison. Continual exploration strategies which use curiosity~\citep{pathak2017curiosity} can be added as an intrinsic reward in the face of non-stationary environments in infinite horizon MDPs~\citep{steinparz2022reactive}. The disagreement between an ensemble of models that predict the next state from the current state and action has been shown to be effective for exploration~\citep{pathak2019self}. Our proposed model uses Plan2Explore which uses forward prediction disagreement and outperforms curiosity-based methods \citep{sekar2020planning}. The tasks themselves can be meta-learned using a latent variable world model and task similarities can be exploited when learning a new task~\citep{fu2022modelbased}.
% \maciej{Strong claim, I think there might be methods for RL adaptation which deal with large shifts} \sam{Honestly I don't think so. This is an area of research I've been thinking about. CRL via quick adaptation. This has been done in CL, not in CRL. Most of Meta-RL deals with 'in-distribution' test tasks.}\maciej{I agree that in general meta-RL algorithms deal with in-distribution stuff. I'd be worried about some outliers like the recent "Human-Timescale Adaptation" paper from DeepMind.}

\textbf{Curriculum Learning}. Another related area of research is open-ended learning which aims to build agents that generalize to unseen environments through a curriculum that starts off with easy tasks and then progresses to harder tasks thereby creating agents which can generalize~\citep{wang2019paired, team2021open, parker2022evolving}.

\section{World Models for Continual Reinforcement Learning}

We leverage world models for learning tasks sequentially without forgetting. We use DreamerV2~\citep{hafner2020mastering} which introduces a discrete stochastic and recurrent world model that is state of the art on numerous single-GPU RL benchmarks. This is a good choice for CRL since the world model is trained by reconstructing state, action, and reward trajectories from experience, we can thus leverage experience replay buffers which persist across tasks to prevent \emph{forgetting} in the world model. Additionally, we can train a policy in the imagination or in the generated trajectories of the world model, similar to generative experience replay methods in supervised CL which remember previous tasks by replaying generated data~\citep{Shin2017}. Thus, using a world model is also sample efficient. Also, world models are \emph{task-agnostic} and do not require external supervision about task changes, without signaling to the agent that it is interacting with a new task. Additionally, by generating rollouts in the world model's imagination the uncertainty in the world model's predictions, more specifically the disagreement between predictions can be used as a task-agnostic exploration bonus. To summarize, we propose using model-based RL with recurrent world models as a viable method for CRL, see~\cref{alg:continual-dv2} for an overview, with DreamerV2 as the world model. Recently, world models have been shown to collect diamonds in Minecraft, a very hard skill to achieve which requires the composition of many other skills with DreamerV3~\citep{hafner2023mastering}, the ideas introduced in this manuscript are directly applicable to newer world model methods as well.

\begin{wrapfigure}{r}{0.55\textwidth}
\vspace{-0.3cm}
\centering
\begin{minipage}{0.55\textwidth}
\begin{algorithm}[H]
   \caption{CRL with World Models}
   \label{alg:continual-dv2}
    \begin{algorithmic}[1]
   \STATE {\bfseries Input:} Tasks (environments) $\mathcal{T}_{1:T}$, world model $M$, policy $\pi$, experience replay buffer $\mathcal{D}$.   
   \FOR{$\mathcal{T}_1$ {\bfseries to} $\mathcal{T}_T$}
   \STATE Train world model $M$ on $\mathcal{D}$.
   \STATE Train $\pi$ inside world model $M$.
   \STATE Execute $\pi$ in task $\mathcal{T}_{\tau}$ to gather episodes and append to $\mathcal{D}$.
   \ENDFOR
\end{algorithmic}
\end{algorithm}
\end{minipage}
\vspace{0.1cm}
\end{wrapfigure}

\textbf{Learning the World Model.} DreamerV2 learns a recurrent (latent) state-space world model (RSSM) which predicts the forward dynamics of the environment. At each time step $t$ the world model receives an observation $o_t$ and is required to reconstruct the observations, $o_t$ conditioned on the previous actions $a_{<t}$ (in addition to reconstructing rewards and discounts). The forward dynamics are modeled using an RNN, $h_t = \textrm{GRU}(h_{t-1}, z_t, a_t)$~\citep{chung2014empirical} where $h_t$ is the hidden state $z_t$ are the discrete probabilistic latent states~\citep{van2017neural} which condition the observation predictions $p(o_t | z_t, h_t)$. Trajectories are sampled from an experience replay buffer and so persisting the replay buffer across different tasks should alleviate forgetting in the world model~\citep{clear}.

\textbf{Policy Learning inside the World Model.} The policy $\pi$ is learned inside the world model by using an actor-critic~\citep{sutton2018reinforcement} while freezing the weights of the RSSM world model. At each step $t$ of the dream inside the RSSM world model a latent state $z_t$ is sampled, $z_t$, and the RNN hidden state condition the actor $\hat{a}_t \sim \pi( \, \cdot \,| z_t, h_t)$. The reward $\hat{r}_{t+1}$ is predicted by the world model. The policy, $\pi$ is then used to obtain new trajectories in the real environment. These trajectories are added to the experience replay buffer. An initial observation $o_1$ is used to start  generating rollouts for policy learning. This training regime ensures that the policy generalizes to previously seen environments through the world model.

\textbf{Task-agnostic Exploration.} The policy learns using the imagined trajectories from the RSSM world model and the world model's predicted rewards are used as a signal for the agent's policy and critic. The policy is also used to gain experience inside the real environment. For exploration, the policy prioritizes regions of the state and action space where the world model produces uncertain predictions. Hence, the uncertainty in the world model's trajectory prediction can be used as an additional intrinsic reward. This idea underpins Plan2Explore \citep{sekar2020planning} which naturally fits with DreamerV2. 

The world model quantifies the uncertainty in the next latent state prediction by using a deep ensemble; multiple neural networks with independent weights. Deep ensembles are a surprisingly robust baseline for uncertainty quantification \citep{lakshminarayanan2017simple} and the ensemble's variance is used as an intrinsic reward. The exploration neural networks in the ensemble are trained to predict the next RSSM latent features $[z_{t+1}, h_{t+1}]$. The world model is frozen while the ensemble is trained. 

The policy $\pi$ observes the reward $r = \alpha_i r_i + \alpha_e r_e$, where $r_e$ is the extrinsic reward predicted by the world model, $r_i$ is the intrinsic reward, the latent disagreement between the next latent state predictions. The coefficients $\alpha_i$ and $\alpha_e$ are $\in [0, 1]$. Hence the policy $\pi$ can be trained inside the world model to seek regions in the state-action space that the world model struggles to predict and hence when the policy is deployed in the environment it will seek these same regions in the state-action space to obtain new trajectories to train the RSSM world model. The exploration strategy is significant for CRL since it is not task dependent unlike using DQN where each task needs an $\epsilon$-greedy schedule~\citep{ewc, kessler2021same} or SAC~\citep{haarnoja2018soft} which needs an entropy regularizer per task~\citep{wolczyk2021continual}.

\subsection{Selective experience replay methods}
\label{sec:selective_exp_replay}

To enable Continual-Dreamer to remember how to solve all the previous tasks it has learned with a limited replay buffer size requires us to select important trajectories to fill the experience replay buffer and selectively choose trajectories to train the world model. DreamerV2 uses a first-in-first-out (FIFO) replay buffer. It also randomly samples trajectories from the replay buffer to train the world model on and also randomly samples a trajectory so that the world model can start dreaming and allow the policy can learn inside the dream. In such a scenario, catastrophic forgetting can occur due to the loss of experience from previous tasks since the replay buffer is FIFO. There is prior work on managing experience replay buffers from supervised CL~\citep{caccia2019online} and off-policy RL~\citep{isele2018selective} which we can systematically study for the application of CRL with world models. To ensure that we have a uniform sample of experience from all tasks in the replay buffer to sample from we explore the following methods:

\begin{itemize}
    \item \textbf{Reservoir Sampling} (\textit{rs}) \citep{Vitter1985RandomSW, isele2018selective}: enables a uniform distribution over all task experience seen so far. This is achieved by storing new examples in the replay buffer with a decreasing probability of $\min(n/t, 1)$, where $t$ is the number of trajectories seen so far and $n$ is the size of the replay buffer. This can be paired with any method of sampling from the replay buffer. By default, this is using random sampling. Reservoir sampling does not check for duplicate examples when storing experience. In practice, we found no duplicates when checking the episodes in the replay buffer using the experimental setup in~\cref{sec:results_minigrid}, this is not a surprise since our task environments are procedurally generated, see~\cref{sec:experiments} for further experimental details.
    \item \textbf{Coverage Maximization} (\textit{cm}) \citep{isele2018selective}: also attempts to create a uniform distribution of experience seen so far. Experience is added to the replay buffer by checking how close it is to trajectories already stored in the replay buffer. Trajectories are embedded using a fixed convolutional LSTM architecture~\citep{shi2015convolutional} and we can calculate distances using an $\normltwo$ distance between the LSTM hidden state with respect to $1000$ randomly selected trajectories from the replay buffer. The median distance determines the priority for the sample to be added to the replay buffer.
\end{itemize}

In addition to methods that populate the experience replay buffer, we can also consider how we should construct the mini-batch for world model and policy learning. For instance,  we can prioritize more informative samples to help remembering and help learning new tasks to aid stability and plasticity. We consider $3$ approaches:

\begin{figure*}
    \centering
    \vspace{-0.5cm}
    \includegraphics[width=0.9\textwidth]{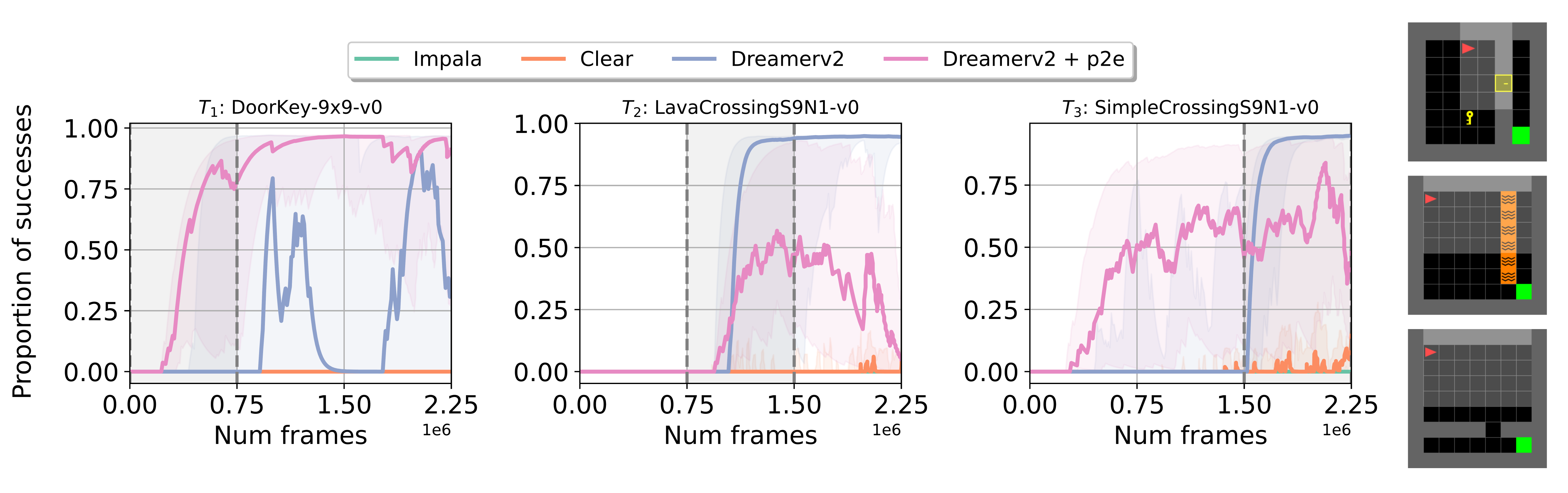}
    \caption{Performance of CRL agents on $3$ Minigrid tasks. Grey-shaded regions indicate the environment which the agent is currently interacting with. All learning curves are a median and inter-quartile range across $20$ seeds. On the right, we pick a random instantiation of the Minigrid environments that are being evaluated. %\piotrm{We do not have Continual-Dreamer here} \sam{No, dreamerv2 + p2e already does well. In the minihack experiments, we stress test this method and require continual-dreamer.}
    }
    \label{fig:minigrid_cl}
    \vspace{-0.5cm}
\end{figure*}

\begin{itemize}
    \item \textbf{Uncertainty sampling} (\textit{us}): we construct a minibatch of experience where the probability of sampling an episode corresponds to the trajectory's uncertainty or intrinsic reward from Plan2Explore. Next state uncertainties are generated for each transition and summed and normalized per trajectory before it is added to the replay buffer. We only calculate the uncertainty once before it is added to the replay buffer. This is similar to sampling the replay buffer according to the size of the temporal-difference error, known as sampling via ``surprise''~\citep{isele2018selective}. The temporal difference error is also only calculated once when transitions are added to the experience replay buffer for DQN.
     \item \textbf{Reward sampling} (\textit{rwd}) \citep{isele2018selective}: we construct a mini-batch of experience for world model learning where the probability that an episode is sampled, corresponds to the reward from the environment.
    \item \textbf{$50$:$50$ sampling}, of past and recent experience. We construct a mini-batch for world model learning based on a $50$:$50$ ratio of uniform random sampling from the replay buffer and sampling from a triangular distribution that favors the most recent experience added so far to help learning more recent tasks. This idea is similar to the on-policy off-policy ratio of learning in CLEAR~\citep{clear} which also aims to balance stability and plasticity.
\end{itemize}
\subsection{Task-aware baseline}
All replay buffer management techniques presented above are task-agnostic, i.e.  operate without explicit task identification.  We also consider a task-aware baseline for comparison, we use $L^2$ weight regularization with respect to the weights from the previous task, which is a simple regularization-based approach to CRL. In this scenario, after the first task, we add to each loss function, an additional objective that minimizes the distance between the current model and policy weights and the optimal weights from the previous task.

\section{Experiments}
\label{sec:experiments}
Our results indicate that DreamerV2 and DreamerV2 + Plan2Explore obtain good out-of-the-box performance for CRL on $3$~Minigrid tasks \citep{gym_minigrid}. On a harder Minihack~\citep{samvelyan2021minihack} tasks from the CORA suite \citep{powers2021cora}, we find that DreamerV2 and DreamerV2 + Plan2Explore exhibit forgetting. To address forgetting we systematically study various selective experience replay methods. The best configuration uses reservoir sampling~\citep{Vitter1985RandomSW} which we name Continual-Dreamer and Continual-Dreamer + Plan2Explore.
%\piotrm{I'd be good to have here td;lr of the results, e.g. point to the most important conclusion etc.}
%\piotrm{Continual-Dreamer should appera somewhat in the intro of this section.}
We use two primary baselines. First, Impala which is a powerful deep RL method not designed for CRL \citep{espeholt2018impala}. Second, we consider CLEAR \citep{clear} which uses Impala as a base RL algorithm and leverages experience replay buffers to prevent forgetting and is task-agnostic. 

Throughout our experiments, we use $3$ different metrics average performance, average forgetting, and average forward transfer~\cref{app:metrics}, to assess the effectiveness of each method~\citep{wolczyk2021continual} in addition to qualitatively inspecting the learning curves. %In particular, we measure the \emph{performance}, which measures how well a CRL method performs on all tasks at the end of the task sequence. We also measure \emph{forgetting}, which is the performance difference after interacting with a task versus the performance at the end of the final task. Finally, we also measure the \emph{forward transfer}, which is the difference in task performance during continual learning compared to single task performance. See~\cref{app:metrics} for their definitions.

\begin{wrapfigure}{r}{0.40\textwidth}
    \centering
    \vspace{-1.0cm}
    \includegraphics[width=0.40\textwidth]{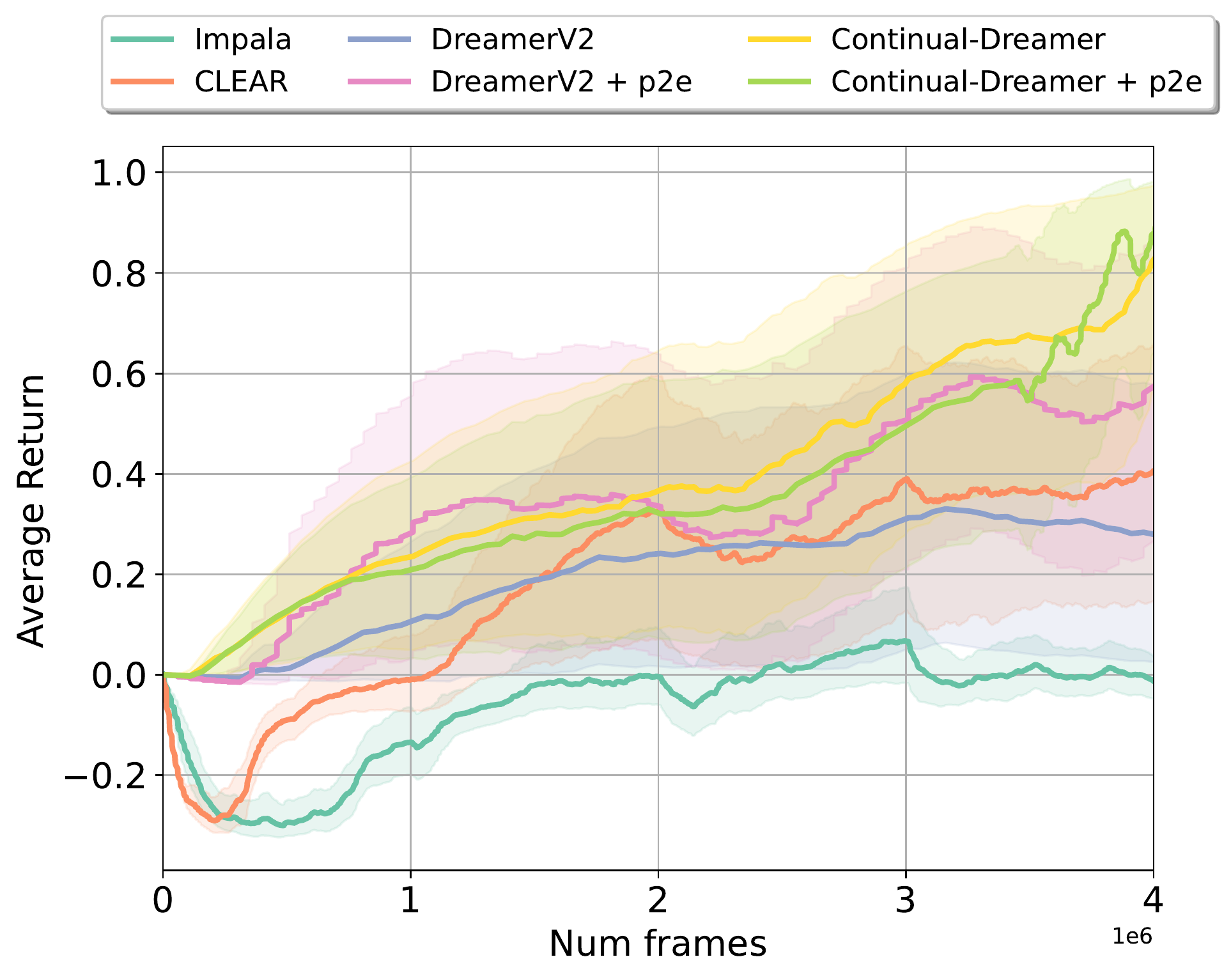}
    \caption{Return averaged over all tasks for various CRL agents on $4$ Minihack tasks. All learning curves are IQM from the \texttt{rliable} package across $10$ seeds and $1000$ bootstrap samples~\citep{agarwal2021deep}.}
    \vspace{-0.2cm}
    \label{fig:minihack_cl}
\end{wrapfigure}

\subsection{Minigrid}
\label{sec:results_minigrid}
\newcommand{\DoorKey}{\texttt{DoorKey-9x9}}
\newcommand{\SimpleCrossing}{\texttt{SimpleCrossing-SN9}}
\newcommand{\LavaCrossing}{\texttt{LavaCrossing-9x9}}

\begin{table*}[]
    \centering
    \resizebox{0.8\textwidth}{!}{\begin{tabular}{lccc}
        \toprule
         & Avg. Performance ($\uparrow$) & Avg. Forgetting ($\downarrow$) & Avg. Forward Transfer ($\uparrow$) \\
         \midrule
         %Continual Option Critic & $0.02 \pm 0.03$ & $0.01 \pm 0.02$ & - \\
         Impala & $0.00 \pm 0.00$ & $0.00 \pm 0.00$ & $0.00 \pm 0.00$ \\ 
         CLEAR & $0.03 \pm 0.05$ & $0.01 \pm 0.06$  & $0.03 \pm 0.03$ \\
         \midrule
         Impala$\times 10$ & $0.16 \pm 0.16$ & $0.06 \pm 0.13$ & - \\
         CLEAR$\times 10$ & $0.64 \pm 0.20$ & $0.00 \pm 0.00$  & - \\
         \midrule
         DreamerV2 & $0.72 \pm 0.24$ & $-0.11 \pm 0.30$ & $0.49 \pm 0.83$ \\ 
         DreamerV2 + p2e & $0.46 \pm 0.10$ & $0.05 \pm 0.18$ & $0.43 \pm 0.22$ \\
         \bottomrule
    \end{tabular}
    }
    \caption{Results on $3$ Minigrid tasks. All metrics are an average and standard deviation over $20$ seeds. We use $0.75$M interactions for each task and $7.5$M in methods marked with $\times 10$. $\uparrow$ indicates better performance with higher numbers, and $\downarrow$ the opposite.}
    \vspace{-0.5cm}
    \label{tab:minigrid}
\end{table*}

We test the out-of-the-box performance of DreamerV2 and DreamerV2 + Plan2Explore as a CRL baseline on $3$ sequential Minigrid tasks. Minigrid is a challenging image-based, partially observable, and sparse reward environment. The agent, in red, will get a reward of $+1$ when it gets to the green goal~\cref{fig:minigrid_cl}. The agent sees a small region of the Minigrid environment as observation, $o_t$. We use $3$ different tasks from Minigrid: \DoorKey{}, \SimpleCrossing{} and \LavaCrossing{}. Each environment has a different skill and so the tasks are diverse. Each method interacts with each task for $0.75$M environment interactions, as previously proposed in \citep{kessler2021same}.

We evaluate CRL agents on all tasks, see~\cref{fig:minigrid_cl}. The results indicate that DreamerV2 is able to solve difficult exploration tasks like the \DoorKey{}.
%which involves the agent having to pick up a key and then use the key to open a door before accessing the goal \piotrm{<-Do we need this story here?} \sam{not really}. 
Additionally, since DreamerV2 trains its policy inside the world model it is more sample efficient than CLEAR which needs $\times10$ more environment interactions to be able to solve the easier Minigrid tasks \SimpleCrossing{} and \LavaCrossing{}, \cref{tab:minigrid}. The addition of Plan2Explore enables DreamerV2 to solve these environments even more quickly, see~\cref{fig:minigrid_cl}. DreamerV2 does exhibit some forgetting of the \DoorKey{} task and this indicates that additional mechanisms to prevent forgetting might be needed.

\begin{figure*}[!ht]
    \centering
    %\vspace{-0.2cm}
    \includegraphics[width=0.95\textwidth]{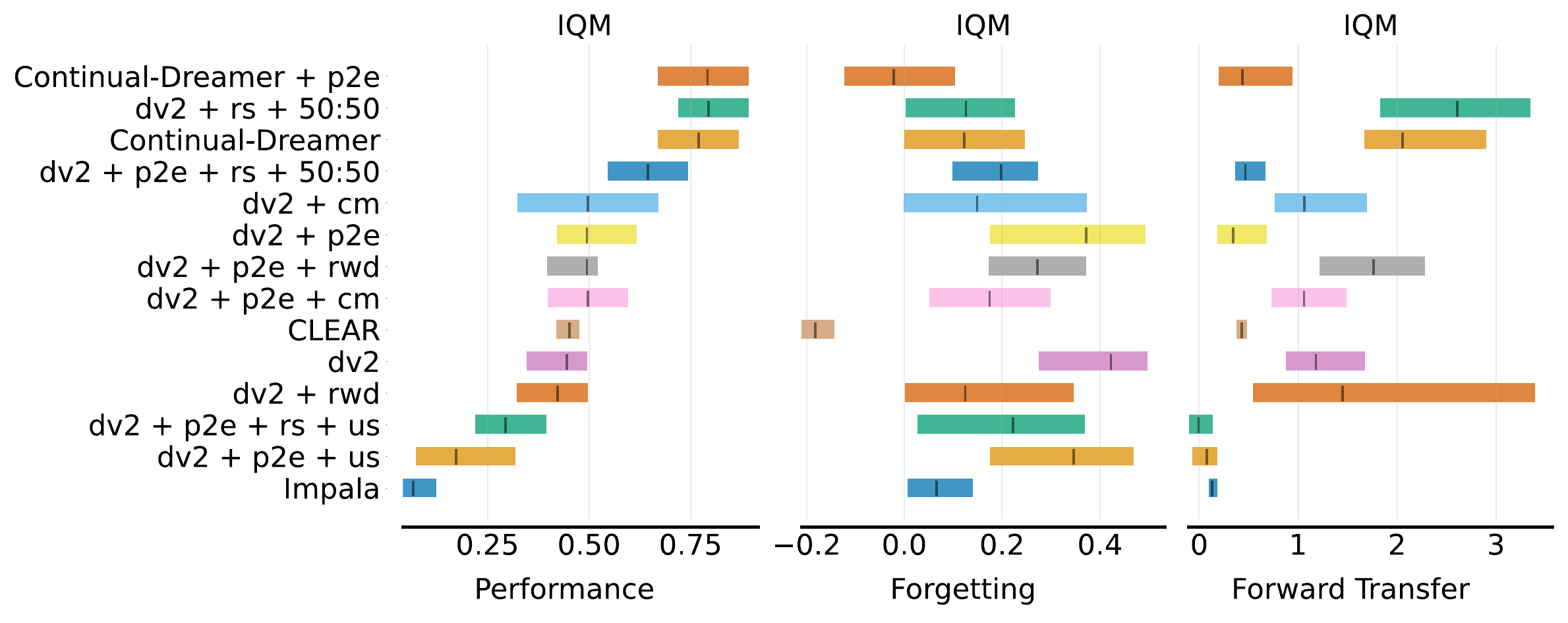}%{plots/4_task_minihack_exp_replay_iqm.drawio.pdf}
    
    % \subfigure{\includegraphics[width=0.45\textwidth]{plots/4_task_minihack_performance.pdf}}
    % \subfigure{\includegraphics[width=0.24\textwidth]{plots/4_task_minihack_forgetting.pdf}}
    % \subfigure{\includegraphics[width=0.225\textwidth]{plots/4_task_minihack_fw_trans.pdf}}
    \caption{Metrics on $4$ Minihack tasks using interquartile mean with $20$ runs with different seeds and $1000$ bootstrap samples from the \texttt{rliable} package~\citep{agarwal2021deep}.}
    \vspace{-0.0cm}
    \label{fig:minihack_exp_replay}
\end{figure*}

From the metrics in \cref{tab:minigrid} we can see that DreamerV2 has strong forward transfer. From the learning curves for individual tasks~\cref{fig:dv2_minigrid_single} we can see that DreamerV2 struggles with independent task learning over the course of $1$M environment steps. In contrast, when learning continually DreamerV2 is able to solve all tasks indicating that it transfers knowledge from previous tasks. This is not entirely surprising since the levels look similar and so the world model will be able to reconstruct observations of a new task more quickly compared to reconstruction from scratch.

For DreamerV2 we use the model and hyperparameters from \citep{hafner2020mastering} with an experience replay buffer for world model learning of size $2$M. For DreamerV2 + Plan2Explore we set the reward coefficients to $\alpha_i = \alpha_e = 0.9$ which was found by grid search of various single task Minihack environments over $ \{0.1, 0.5, 0.9\}$ we use the same policy for exploration and evaluation and learn the world model by observation reconstruction only, rather than observation, reward, and discount reconstruction. We explore these design decisions using the Minihack benchmark in~\Cref{sec:minihack_ablations}. For CLEAR we also use an experience replay buffer size of $2$M like DreamerV2 and DreamerV2 + Plan2Explore.
\looseness=-1
\subsection{Minihack}
\label{sec:results_minihack}

\newcommand{\RoomRandom}{\texttt{Room-Random-15x15-v0}}
\newcommand{\RoomMonster}{\texttt{Room-Monster-15x15-v0}}
\newcommand{\RoomTrap}{\texttt{Room-Trap-15x15-v0}}
\newcommand{\RoomUltimate}{\texttt{Room-Ultimate-15x15-v0}}
\newcommand{\RiverNarrow}{\texttt{River-Narrow-v0}}
\newcommand{\River}{\texttt{River-v0}}
\newcommand{\RiverMonster}{\texttt{River-Monster-v0}}
\newcommand{\HideNSeek}{\texttt{HideNSeek-v0}}
\newcommand{\FourRooms}{\texttt{FourRooms-v0}}

%\sam{say world models rock first. i.e. solving \RiverMonster{}.}\piotrm{why?} \sam{not required.}

\begin{wrapfigure}{r}{0.45\textwidth}
    \centering
    \vspace{-0.1cm}
    \includegraphics[width=0.45\textwidth]{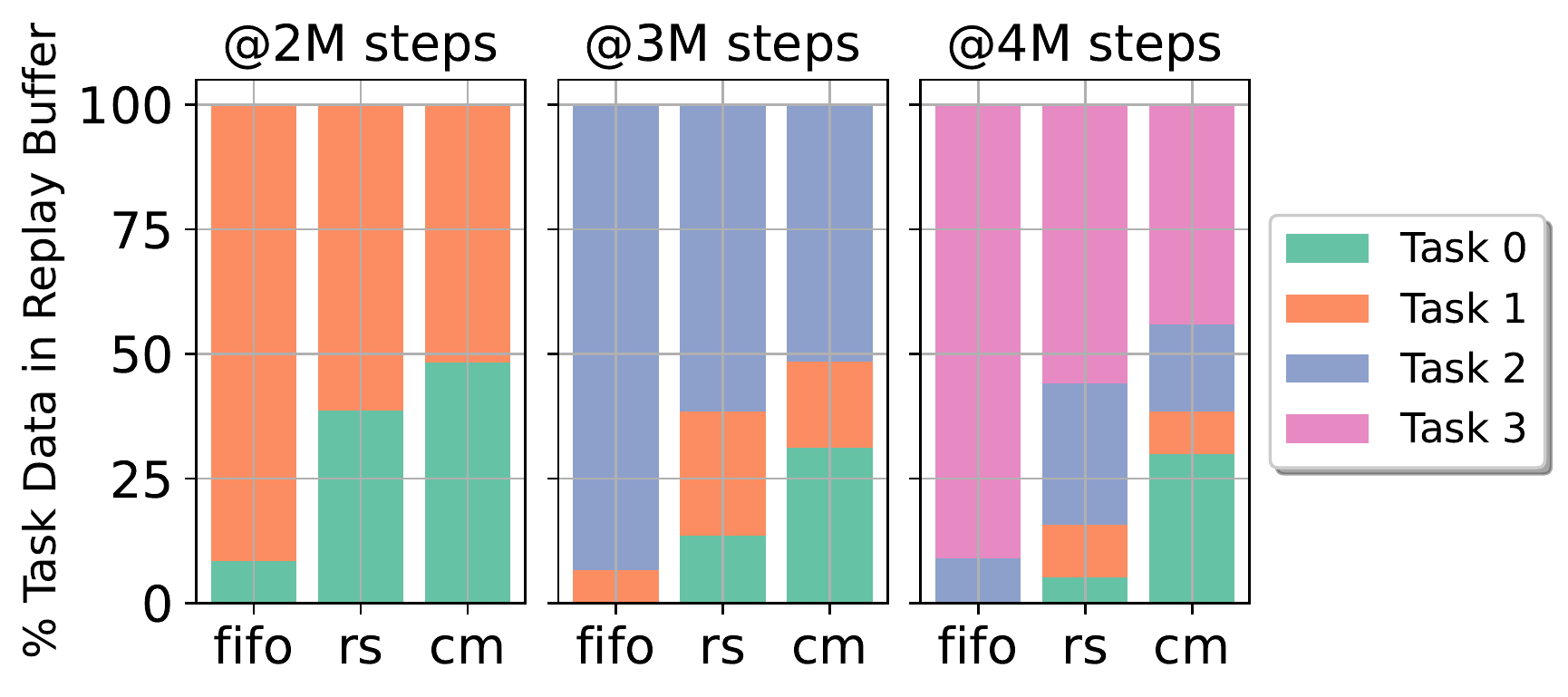}
    \caption{Proportion of episodes from each task in the replay buffer for different replay buffer construction strategies at $2$M, $3$M, and $4$M environment steps on $4$ task Minihack. Bar plots are an average of $5$ runs.}
    \vspace{-0.5cm}
    \label{fig:task_distribution_cl_small}
\end{wrapfigure}

We test DreamerV2 and DreamerV2 with Plan2Explore on a set of harder Minihack tasks~\citep{samvelyan2021minihack}.  Minihack is a set of diverse, image-based, and sparse reward tasks based on the game of Nethack \citep{kuettler2020nethack} which have larger state spaces than MiniGrid and require learning more difficult skills such as crossing rivers by pushing multiple rocks into the river for instance. This will test the task-agnostic exploration mechanism from Plan2Explore further. We use $4$ tasks Minihack tasks: in particular, we consider the following tasks \RoomRandom{}, \RoomTrap{}, \RiverNarrow{}, and \RiverMonster{}  which are a subset of the $12$ Minihack tasks from the CORA CRL benchmark \citep{powers2021cora}. Each task is seen once and has a budget of $1$M environment interactions. We use the same experimental setup as in~\cref{sec:results_minigrid}, however, we keep the sizes of the experience replay buffer fixed to $1$M for both DreamerV2, its variants, and CLEAR.

\looseness=-1 We perform a comprehensive evaluation of various replay buffer management and mini-batch selection strategies. We find that using reservoir sampling together with DreamerV2 and DreamerV2 + Plan2Explore is the best configuration which we name Continual-Dreamer, see detailed analysis in Section \ref{sec:selective_replay_methods}. The results of the Minihack experiment, as shown in ~\cref{fig:minihack_cl}, demonstrate that Continual-Dreamer and Continual-Dreamer + Plan2Explore perform better than the baselines regarding the average return over all tasks and $10$ seeds.  In particular, Continual-Dreamer and Continual-Dreamer + Plan2Explore exhibit faster learning on new tasks. Neither task-agnostic continual learning baselines, Impala and CLEAR, can learn the most difficult task \RiverMonster{},~\cref{fig:app:mh_4_task_return}. These results suggest that world models are effective for consolidating knowledge across environments and changing task objectives. We also compare to a task-aware baseline: an $\normltwo$ regularization of the world model and actor-critic about the previous task's optimal parameters~\cref{fig:app:task_aware}. We find that this performs poorly.

\subsubsection{Evaluation of Different Selective Replay Methods} \label{sec:selective_replay_methods}

\cref{fig:minihack_exp_replay} presents results for different replay strategies. If we consider DreamerV2 and DreamerV2 + Plan2Explore we can see that there is some forgetting of the first Room tasks, see~\cref{fig:app:mh_4_task_return}. Our main finding is that reservoir sampling robustly mitigates forgetting, which gives rise to Continual-Dreamer. 

We can increase plasticity by biasing the sampling toward recent experiences. We can see that if we add $50$:$50$ sampling of the minibatch construction together with reservoir sampling, this causes inconsistent effects. For Continual-Dreamer, $50$:$50$ sampling increases forgetting while the performance remains constant, indicating better learning of harder tasks and transfer. On the other hand, when $50$:$50$ sampling is applied to Continual-Dreamer + Plan2Explore performance and forgetting worsen. 

We tested DreamerV2's performance in other variants, including uncertainty sampling (\textit{us}), coverage maximization (\textit{cm}), and reward sampling (\textit{rwd}). The results we obtained are consistent with prior works~\citep{isele2018selective}. It can be seen that \textit{us} performed closely to Impala with low performance, forward transfer, and high forgetting. Using \textit{rwd} sampling results in performance that does not improve over random sampling of the minibatch. Using \textit{cm} with DreamerV2 and DreamerV2 + Plan2Explore results in less forgetting. However, it behaves inconsistently when observing performance; performance increases when applied to DreamerV2 and it decreases when applied to DreamerV2 + Plan2Explore. As a baseline, we also tested a naive approach, which is to increase the size of the replay buffer; this indeed prevents forgetting and increases performance, see~\Cref{sec:minihack_buffer_sz}. However, this is at the cost of making it harder to learn new tasks: the harder exploration Minihack tasks are not solved and forward transfer decreases as we increase the size of the replay buffer. In~\cref{fig:task_distribution_cl_small} we can see the mini-batch task composition at $2$M, $3$M and $4$M steps for $4$ task Minihack for various replay buffer management methods. As training progresses the FIFO will lose experience from earlier tasks, whereas \emph{rs} and \emph{cm} are able to retain experience from earlier tasks. \emph{rs} retains a more uniform distribution. Whereas \emph{cm} - which is distance based - retains fewer samples from intermediate tasks, e.g. Task 1 at $4$M steps since embeddings look similar to Task $0$. We store entire episodes of state, action, and rewards in the replay buffer. When a task is mastered shorter episodes where the agent navigates to the goal are stored in the replay buffer. When adding  experience from a new task the episodes are longer (episodes can reach the cut-off of $100$ transitions before the episode is ended) and thus require more episodes of past data to be removed to make way for it in the reservoir sampling update. Due to this phenomenon, there are relatively fewer samples from the earlier tasks in the replay buffer. We perform a similar analysis for the sampling methods we consider in~\cref{sec:sampling_distribution}.

\begin{figure}
    \centering
    \includegraphics[width=1.0\textwidth]{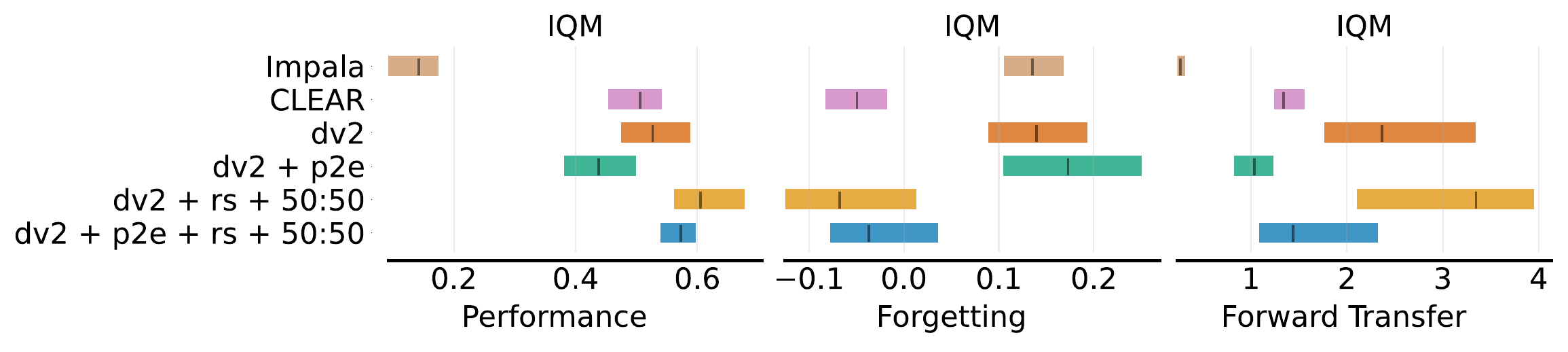}
    \caption{Metrics on $8$ Minihack tasks using interquartile mean with $10$ runs with different seeds and $1000$ bootstrap samples. The results for DreamerV2 + rs + $50$:$50$ with and without Plan2Explore are an interquartile mean with $5$ runs with different seeds and $1000$ bootstrap samples.}
    \label{fig:8task_iqm}
    \vspace{-0.7cm}
\end{figure}

\subsection{Scaling to More Tasks}

We scale to $8$ Minihack tasks from the CORA suite: \RoomRandom{}, \RoomMonster{}, \RoomTrap{}, \RoomUltimate{}, \RiverNarrow{}, \River{}, \RiverMonster{} and \HideNSeek{}. For DreamerV2 and its variants and for CLEAR we set the size of the experience replay buffer to $2$M. We can see that by using reservoir sampling with $50$:$50$ sampling prevents forgetting and slightly improves performance in DreamerV2 and DreamerV2 + Plan2Explore~\cref{fig:8task_iqm}. Performance over CLEAR is also improved by introducing reservoir sampling and $50$:$50$ sampling. The difficult Minihack tasks, such as \River{}, \RiverMonster{} and \HideNSeek{} are solved by DreamerV2 and its variants but not by CLEAR~\cref{fig:app:8task_mh}. We conjecture that warm-starting the world model by solving easier related tasks enables DreamerV2 to improve its downstream performance. For instance the single-task DreamerV2, see ~\cref{fig:dv2_minihack_single}, does not solve \RiverMonster{} as often as continual agent~\cref{fig:app:8task_mh}.
%For instance the single task \RiverMonster{} task learning curves~\cref{fig:dv2_minihack_single} show that DreamerV2 and its variants do not solve these tasks as often as the continual agent~\cref{fig:app:8task_mh}. 
Arguably, the skills learned in \RoomMonster{} and \River{} enable DreamerV2 to solve \RiverMonster{}.

\begin{comment}
    
\subsection{Where is the Source of Forgetting?}

 \begin{wrapfigure}{r}{0.3\textwidth}
    \centering
    \vspace{-0.5cm}
    \includegraphics[width=0.3\textwidth]{plots/sep_ac_iqm.drawio.pdf}
    \caption{Metrics on $4$ Minihack tasks using interquartile mean with $5$ runs with different seeds and $1000$ bootstrap samples from DreamerV2 versus DreamerV2 with a separate actor and critic per task.}
    \label{fig:sep_ac_iqm}
    \vspace{-0.4cm}
\end{wrapfigure}
\piotrm{I opt to remove this section. What is says, is that if we train the world model and do not train AC then things break, which is obvious and does not yield any useful conclusion?}
\sam{Really? it is obvious in hindsight.}

To determine the source of forgetting, we test an augmented DreamerV2 with a separate actor-critic for each task. When the task changes, we initiate a new actor-critic and freeze the previous one. Task information is necessary for evaluation. We can see from the IQM metrics in~\cref{fig:sep_ac_iqm} that performance is lower and forgetting is higher than just using a single actor-critic in a task-agnostic fashion. This shows that the most important component is the world model in the CRL setup. Since the world model representations are constantly changing the previous task frozen actor-critics are unable to use them to select appropriate actions. This motivates us to develop CL methods that focus on the world model.
\end{comment}

\section{Limitations}

\textbf{Interference}. Continual-Dreamer uses a replay buffer of past experience for retaining skills. Interference from past data is when past samples prevent the learning of new tasks in the face of changing reward functions or goal locations. This has been shown in world-models~\citep{wan2022towards} and off-policy RL~\citep{kessler2021same}. Continual-Dreamer and the selective experience replay methods which are explored are not intended to prevent interference from past samples.
We show interference can affect DreamerV2 on the Minigrid \FourRooms{} environment with changing rewards/goals for $2$ different tasks in~\cref{sec:4R_interfernce}.
%We observe this interference explicitly using DreamerV2 with Plan2Explore on the Minigrid \FourRooms{} environment where we change the reward function or goal location from one task to another~\cref{fig:interference}. We train for $1$M steps on the first task, then $1$M steps on the second task where the goal location has changed with all else the same. From the learning curves for each individual run, we can see how only one task will ever be solved with simultaneously poor performance in the second task (with the exception of one seed, in blue, which switches between solving and not solving each task).} \piotrm{I would probably make this paragraph shorter, stating that 'we do not tackle interference and defer it to a further study' and put the example to the appendix. But this is probably minor.} \sam{all reviewers were pretty keen we discuss this, also given we put this in the rebuttal, maybe the reviewers expect this in the revise and resubmit.}

\begin{comment}
\begin{wrapfigure}{r}{0.45\textwidth}
    \centering
    \vspace{-0.8cm}
    \includegraphics[width=0.45\textwidth]{plots/interference.drawio.pdf}
    \caption{\textcolor{orange}{\textbf{Top}, $2$ \FourRooms{} environment on with fixed agent start location, agent start orientation, and obstacles. Only the goal or reward function changes from one task to the next. \textbf{Bottom}, $4$ separate success rate learning curves for different random seeds in different colours for DreamerV2 + p2e.}}
    \vspace{-0.5cm}
    \label{fig:interference}
\end{wrapfigure}
\end{comment}

\begin{wrapfigure}{r}{0.45\textwidth}
    \centering
    \vspace{-1.0cm}
    \includegraphics[width=0.45\textwidth]{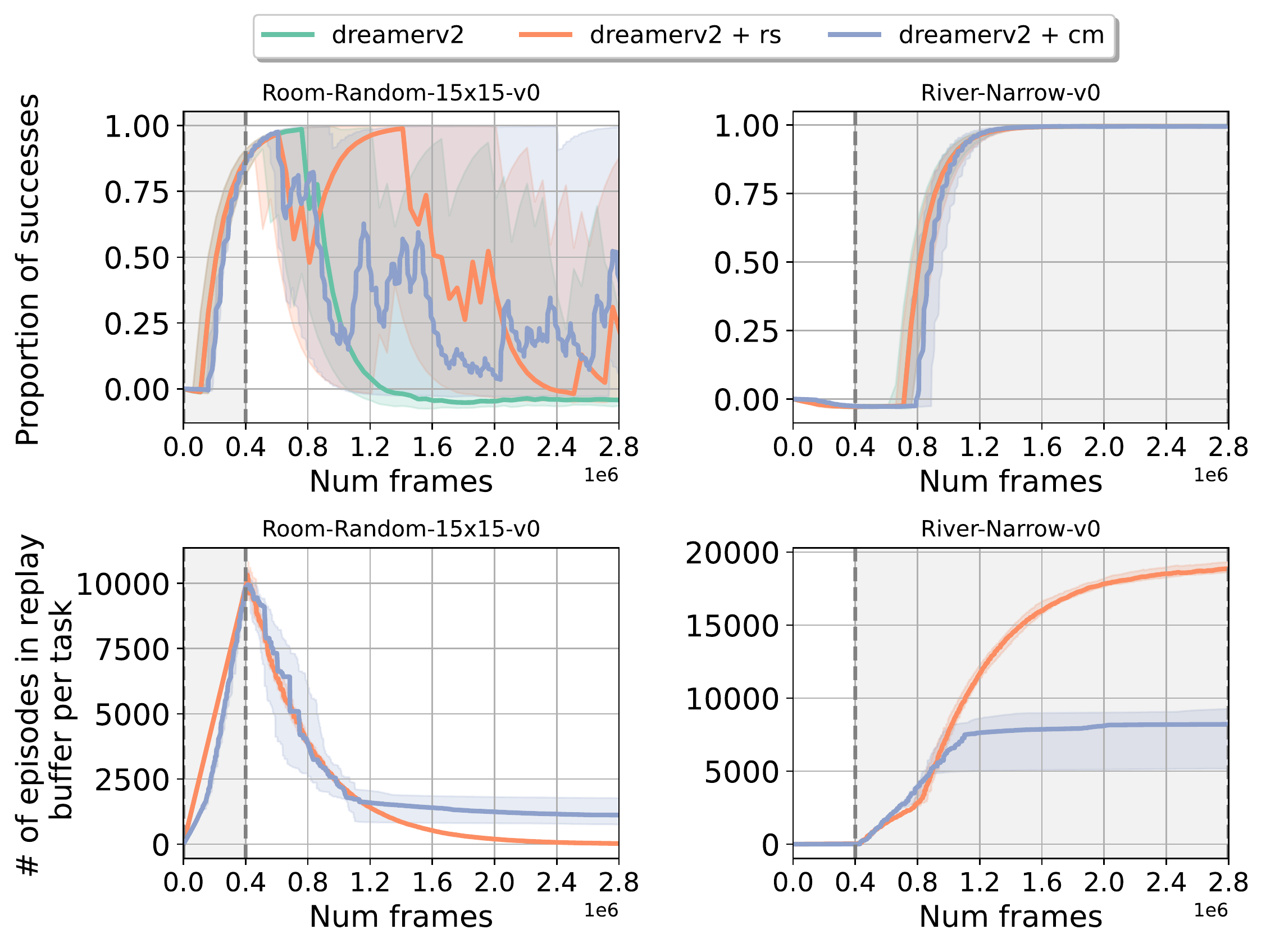}
    \caption{\textbf{Top}, learning curves for imbalanced number of interactions with \RoomRandom{} and \RiverNarrow{}. \textbf{Bottom}, the number of episodes from each task in the replay buffer for \emph{rs} and \emph{cm}. The grey shaded region indicates which timesteps the agent interacts with a task. All runs are interquartile ranges over $20$ random seeds.}
    \vspace{-0.0cm}
    \label{fig:imbalance}
\end{wrapfigure}

\textbf{Task data imbalances.} The selective experience replay considered for managing experience in the replay buffer are not designed for keeping an even distribution of task data in the replay buffer in the face of imbalanced tasks. To illustrate this, we consider a $2$ task scenario with $0.4$M interactions of \RoomRandom{} and then $2.4$M interactions of \RiverNarrow{} and a replay buffer of size $0.4$M. We see that both \emph{rs} and \emph{cm} are unable to retain the performance of the first \RoomRandom{} similarly to a FIFO buffer~\cref{fig:imbalance}. For \emph{rs} experience from the longer second task will saturate the replay buffer. While \emph{cm} is able to retain a small number of samples from the first task, but not enough to prevent forgetting, since it uses a distanced-based criterion to populate the replay buffer. So \emph{cm}-type approaches to replay buffer management could be a fruitful method for managing task imbalances.

\section{Discussion and Future Works}

We have explored the use of world models as a \emph{task-agnostic} CRL baseline. 
World models can be powerful CRL agents as they train the policy inside the world model and can thus be sample efficient. World models are trained by using experience replay buffers and so we can prevent \emph{forgetting} of past tasks by persisting the replay buffer across  from the current task to new tasks. Importantly, the world model's prediction uncertainty can be used as an additional intrinsic task-agnostic reward to help exploration and solve difficult tasks in a task-agnostic fashion \citep{sekar2020planning}. Previous CRL exploration strategies in the literature all require an indication of when the agent stops interacting with a previous task and starts interacting with a new task to reset an exploration schedule. Our implementation uses DreamerV2 as the world model \citep{hafner2020mastering}, and we demonstrate a selective experience replay setting which we call Continual-Dreamer which is a powerful CRL method on two difficult CRL benchmarks.

We show empirically that world models can be a strong task-agnostic baseline for CRL problems compared to state-of-the-art task-agnostic methods~\citep{clear}. DreamerV2 with Plan2Explore outperforms CLEAR on Minigrid. Our experiments on Minihack test the limits of using world models for CRL and require us to introduce experience replay buffer management methods to aid in retaining skills in addition to enabling the learning of new skills. We show that reservoir sampling enables an even coverage of experience in the replay buffer to mitigate forgetting. We call this configuration of DreamerV2 with reservoir sampling Continual-Dreamer. Future work will explore continuous control CRL benchmarks, such as ContinualWorld~\citep{wolczyk2021continual} and explore other world-models.

\section{Acknowledgements}
We would like to thank anonymous reviewers for their valuable feedback. SK would like to thank the Oxford-Man Institute of Quantitative Finance for funding. This research was funded by National Science Center Poland under the grant agreement 2020/39/B/ST6/01511 and from Warsaw University of Technology within the Excellence Initiative: Research University (IDUB) programme. We gratefully acknowledge Polish high-performance computing infrastructure PLGrid (HPC Centers: ACK Cyfronet AGH, CI TASK) for providing computer facilities and support within computational grant no. PLG/2023/016202. Piotr Milos was supported by the Polish National Science Centre grant 2019/35/O/ST6/03464.

%\bibliography{main}

\begin{thebibliography}{70}
\providecommand{\natexlab}[1]{#1}
\providecommand{\url}[1]{\texttt{#1}}
\expandafter\ifx\csname urlstyle\endcsname\relax
  \providecommand{\doi}[1]{doi: #1}\else
  \providecommand{\doi}{doi: \begingroup \urlstyle{rm}\Url}\fi

\bibitem[Agarwal et~al.(2021)Agarwal, Schwarzer, Castro, Courville, and
  Bellemare]{agarwal2021deep}
Rishabh Agarwal, Max Schwarzer, Pablo~Samuel Castro, Aaron~C Courville, and
  Marc Bellemare.
\newblock Deep reinforcement learning at the edge of the statistical precipice.
\newblock \emph{Advances in neural information processing systems},
  34:\penalty0 29304--29320, 2021.

\bibitem[Aljundi et~al.(2019)Aljundi, Lin, Goujaud, and Bengio]{Aljundi2019}
Rahaf Aljundi, Min Lin, Baptiste Goujaud, and Yoshua Bengio.
\newblock {Gradient based sample selection for online continual learning}.
\newblock In \emph{Advances in Neural Information Processing Systems}, 2019.

\bibitem[Barreto et~al.(2017)Barreto, Dabney, Munos, Hunt, Schaul, van Hasselt,
  and Silver]{barreto2017successor}
Andr{\'e} Barreto, Will Dabney, R{\'e}mi Munos, Jonathan~J Hunt, Tom Schaul,
  Hado~P van Hasselt, and David Silver.
\newblock Successor features for transfer in reinforcement learning.
\newblock \emph{Advances in neural information processing systems}, 30, 2017.

\bibitem[Barreto et~al.(2019)Barreto, Borsa, Hou, Comanici, Ayg{\"u}n, Hamel,
  Toyama, Hunt, Mourad, Silver, et~al.]{barreto2019option}
Andr{\'e} Barreto, Diana Borsa, Shaobo Hou, Gheorghe Comanici, Eser Ayg{\"u}n,
  Philippe Hamel, Daniel~K Toyama, Jonathan~J Hunt, Shibl Mourad, David Silver,
  et~al.
\newblock The option keyboard: Combining skills in reinforcement learning.
\newblock 2019.

\bibitem[Benjamin et~al.(2019)Benjamin, Rolnick, and Kording]{Benjamin2019}
Ari~S Benjamin, David Rolnick, and Konrad~P Kording.
\newblock {Measuring and Regularizing Networks in Function Space}.
\newblock In \emph{International Conference on Learning Representations}, 2019.

\bibitem[Buzzega et~al.(2020)Buzzega, Boschini, Porrello, Abati, and
  Calderara]{buzzega2020dark}
Pietro Buzzega, Matteo Boschini, Angelo Porrello, Davide Abati, and Simone
  Calderara.
\newblock Dark experience for general continual learning: a strong, simple
  baseline.
\newblock \emph{Advances in neural information processing systems},
  33:\penalty0 15920--15930, 2020.

\bibitem[Caccia et~al.(2019)Caccia, Aljundi, Belilovsky, Caccia, Charlin, and
  Tuytelaars]{caccia2019online}
Lucas Caccia, Rahaf Aljundi, Eugene Belilovsky, Massimo Caccia, Laurent
  Charlin, and Tinne Tuytelaars.
\newblock Online continual learning with maximally interfered retrieval.
\newblock \emph{Advances in Neural Information Processing (NeurIPS)}, 2019.

\bibitem[Caccia et~al.(2022)Caccia, Mueller, Kim, Charlin, and
  Fakoor]{caccia2022task}
Massimo Caccia, Jonas Mueller, Taesup Kim, Laurent Charlin, and Rasool Fakoor.
\newblock Task-agnostic continual reinforcement learning: In praise of a simple
  baseline.
\newblock \emph{arXiv preprint arXiv:2205.14495}, 2022.

\bibitem[Chaudhry et~al.(2019)Chaudhry, Facebook, Research, Elhoseiny,
  Ajanthan, Dokania, Torr, and Ranzato]{ChaudhryTinyEps}
Arslan Chaudhry, Marcus~Rohrbach Facebook, A~I Research, Mohamed Elhoseiny,
  Thalaiyasingam Ajanthan, Puneet~K Dokania, Philip H~S Torr, and Marc
  '~Aurelio Ranzato.
\newblock {On Tiny Episodic Memories in Continual Learning}.
\newblock \emph{arxiv.org:1902.10486}, 2019.

\bibitem[Chevalier-Boisvert et~al.(2018)Chevalier-Boisvert, Willems, and
  Pal]{gym_minigrid}
Maxime Chevalier-Boisvert, Lucas Willems, and Suman Pal.
\newblock Minimalistic gridworld environment for openai gym.
\newblock \url{https://github.com/maximecb/gym-minigrid}, 2018.

\bibitem[Chung et~al.(2014)Chung, Gulcehre, Cho, and
  Bengio]{chung2014empirical}
Junyoung Chung, Caglar Gulcehre, KyungHyun Cho, and Yoshua Bengio.
\newblock Empirical evaluation of gated recurrent neural networks on sequence
  modeling.
\newblock \emph{arXiv preprint arXiv:1412.3555}, 2014.

\bibitem[Degrave et~al.(2022)Degrave, Felici, Buchli, Neunert, Tracey,
  Carpanese, Ewalds, Hafner, Abdolmaleki, de~Las~Casas,
  et~al.]{degrave2022magnetic}
Jonas Degrave, Federico Felici, Jonas Buchli, Michael Neunert, Brendan Tracey,
  Francesco Carpanese, Timo Ewalds, Roland Hafner, Abbas Abdolmaleki, Diego
  de~Las~Casas, et~al.
\newblock Magnetic control of tokamak plasmas through deep reinforcement
  learning.
\newblock \emph{Nature}, 602\penalty0 (7897):\penalty0 414--419, 2022.

\bibitem[Espeholt et~al.(2018)Espeholt, Soyer, Munos, Simonyan, Mnih, Ward,
  Doron, Firoiu, Harley, Dunning, et~al.]{espeholt2018impala}
Lasse Espeholt, Hubert Soyer, Remi Munos, Karen Simonyan, Volodymir Mnih, Tom
  Ward, Yotam Doron, Vlad Firoiu, Tim Harley, Iain Dunning, et~al.
\newblock {IMPALA}: {S}calable distributed deep-{RL} with importance weighted
  actor-learner architectures.
\newblock In \emph{International Conference on Machine Learning}. 2018.

\bibitem[Fu et~al.(2022)Fu, Yu, Littman, and Konidaris]{fu2022modelbased}
Haotian Fu, Shangqun Yu, Michael Littman, and George Konidaris.
\newblock Model-based lifelong reinforcement learning with bayesian
  exploration.
\newblock In Alice~H. Oh, Alekh Agarwal, Danielle Belgrave, and Kyunghyun Cho
  (eds.), \emph{Advances in Neural Information Processing Systems}, 2022.
\newblock URL \url{https://openreview.net/forum?id=6I3zJn9Slsb}.

\bibitem[Gaya et~al.(2022)Gaya, Doan, Caccia, Soulier, Denoyer, and
  Raileanu]{gaya2022building}
Jean-Baptiste Gaya, Thang Doan, Lucas Caccia, Laure Soulier, Ludovic Denoyer,
  and Roberta Raileanu.
\newblock Building a subspace of policies for scalable continual learning.
\newblock \emph{arXiv preprint arXiv:2211.10445}, 2022.

\bibitem[Ha \& Schmidhuber(2018)Ha and Schmidhuber]{ha2018world}
David Ha and J{\"u}rgen Schmidhuber.
\newblock World models.
\newblock \emph{arXiv preprint arXiv:1803.10122}, 2018.

\bibitem[Haarnoja et~al.(2018)Haarnoja, Zhou, Abbeel, and
  Levine]{haarnoja2018soft}
Tuomas Haarnoja, Aurick Zhou, Pieter Abbeel, and Sergey Levine.
\newblock Soft actor-critic: Off-policy maximum entropy deep reinforcement
  learning with a stochastic actor.
\newblock In \emph{International conference on machine learning}, pp.\
  1861--1870. PMLR, 2018.

\bibitem[Hafner et~al.(2020)Hafner, Lillicrap, Norouzi, and
  Ba]{hafner2020mastering}
Danijar Hafner, Timothy Lillicrap, Mohammad Norouzi, and Jimmy Ba.
\newblock Mastering atari with discrete world models.
\newblock \emph{arXiv preprint arXiv:2010.02193}, 2020.

\bibitem[Hafner et~al.(2023)Hafner, Pasukonis, Ba, and
  Lillicrap]{hafner2023mastering}
Danijar Hafner, Jurgis Pasukonis, Jimmy Ba, and Timothy Lillicrap.
\newblock Mastering diverse domains through world models.
\newblock \emph{arXiv preprint arXiv:2301.04104}, 2023.

\bibitem[Hassabis et~al.(2017)Hassabis, Kumaran, Summerfield, and
  Botvinick]{HASSABIS2017245}
Demis Hassabis, Dharshan Kumaran, Christopher Summerfield, and Matthew
  Botvinick.
\newblock Neuroscience-inspired artificial intelligence.
\newblock \emph{Neuron}, 95\penalty0 (2):\penalty0 245 -- 258, 2017.
\newblock ISSN 0896-6273.
\newblock \doi{https://doi.org/10.1016/j.neuron.2017.06.011}.

\bibitem[Hausknecht \& Stone(2015)Hausknecht and Stone]{hausknecht2015deep}
Matthew Hausknecht and Peter Stone.
\newblock Deep recurrent q-learning for partially observable mdps.
\newblock In \emph{2015 aaai fall symposium series}, 2015.

\bibitem[Hsu et~al.(2018)Hsu, Liu, Ramasamy, and Kira]{hsu2018re}
Yen-Chang Hsu, Yen-Cheng Liu, Anita Ramasamy, and Zsolt Kira.
\newblock {Re-evaluating continual learning scenarios: A categorization and
  case for strong baselines}.
\newblock \emph{arXiv preprint arXiv:1810.12488}, 2018.

\bibitem[Huang et~al.(2021)Huang, Xie, Bharadhwaj, and
  Shkurti]{huang2021continual}
Yizhou Huang, Kevin Xie, Homanga Bharadhwaj, and Florian Shkurti.
\newblock Continual model-based reinforcement learning with hypernetworks.
\newblock In \emph{2021 IEEE International Conference on Robotics and
  Automation (ICRA)}, pp.\  799--805. IEEE, 2021.

\bibitem[Isele \& Cosgun(2018)Isele and Cosgun]{isele2018selective}
David Isele and Akansel Cosgun.
\newblock Selective experience replay for lifelong learning.
\newblock In \emph{Proceedings of the AAAI Conference on Artificial
  Intelligence}, volume~32, 2018.

\bibitem[Kaelbling et~al.(1998)Kaelbling, Littman, and
  Cassandra]{kaelbling1998planning}
Leslie~Pack Kaelbling, Michael~L Littman, and Anthony~R Cassandra.
\newblock Planning and acting in partially observable stochastic domains.
\newblock \emph{Artificial intelligence}, 101\penalty0 (1-2):\penalty0 99--134,
  1998.

\bibitem[Kaiser et~al.(2019)Kaiser, Babaeizadeh, Milos, Osinski, Campbell,
  Czechowski, Erhan, Finn, Kozakowski, Levine, et~al.]{kaiser2019model}
Lukasz Kaiser, Mohammad Babaeizadeh, Piotr Milos, Blazej Osinski, Roy~H
  Campbell, Konrad Czechowski, Dumitru Erhan, Chelsea Finn, Piotr Kozakowski,
  Sergey Levine, et~al.
\newblock Model-based reinforcement learning for atari.
\newblock \emph{arXiv preprint arXiv:1903.00374}, 2019.

\bibitem[Kessler et~al.(2021)Kessler, Parker-Holder, Ball, Zohren, and
  Roberts]{kessler2021same}
Samuel Kessler, Jack Parker-Holder, Philip Ball, Stefan Zohren, and Stephen~J
  Roberts.
\newblock Same state, different task: Continual reinforcement learning without
  interference.
\newblock \emph{arXiv preprint arXiv:2106.02940}, 2021.

\bibitem[Khetarpal et~al.(2020)Khetarpal, Riemer, Rish, and Precup]{crl_review}
Khimya Khetarpal, Matthew Riemer, Irina Rish, and Doina Precup.
\newblock Towards continual reinforcement learning: A review and perspectives.
\newblock \emph{arXiv preprint arXiv:2012.13490}, 2020.

\bibitem[Kirkpatrick et~al.(2016)Kirkpatrick, Pascanu, Rabinowitz, Veness,
  Desjardins, Rusu, Milan, Quan, Ramalho, Grabska{-}Barwinska, Hassabis,
  Clopath, Kumaran, and Hadsell]{ewc}
James Kirkpatrick, Razvan Pascanu, Neil~C. Rabinowitz, Joel Veness, Guillaume
  Desjardins, Andrei~A. Rusu, Kieran Milan, John Quan, Tiago Ramalho, Agnieszka
  Grabska{-}Barwinska, Demis Hassabis, Claudia Clopath, Dharshan Kumaran, and
  Raia Hadsell.
\newblock Overcoming catastrophic forgetting in neural networks.
\newblock \emph{CoRR}, abs/1612.00796, 2016.

\bibitem[K{\"{u}}ttler et~al.(2020)K{\"{u}}ttler, Nardelli, Miller, Raileanu,
  Selvatici, Grefenstette, and Rockt{\"{a}}schel]{kuettler2020nethack}
Heinrich K{\"{u}}ttler, Nantas Nardelli, Alexander~H. Miller, Roberta Raileanu,
  Marco Selvatici, Edward Grefenstette, and Tim Rockt{\"{a}}schel.
\newblock {The NetHack Learning Environment}.
\newblock In \emph{Proceedings of the Conference on Neural Information
  Processing Systems (NeurIPS)}, 2020.

\bibitem[Lakshminarayanan et~al.(2017)Lakshminarayanan, Pritzel, and
  Blundell]{lakshminarayanan2017simple}
Balaji Lakshminarayanan, Alexander Pritzel, and Charles Blundell.
\newblock Simple and scalable predictive uncertainty estimation using deep
  ensembles.
\newblock \emph{Advances in neural information processing systems}, 30, 2017.

\bibitem[Lee et~al.(2020)Lee, Ha, Zhang, and Kim]{lee2020neural}
Soochan Lee, Junsoo Ha, Dongsu Zhang, and Gunhee Kim.
\newblock A neural dirichlet process mixture model for task-free continual
  learning.
\newblock \emph{arXiv preprint arXiv:2001.00689}, 2020.

\bibitem[Li \& Hoiem(2017)Li and Hoiem]{Li2017}
Zhizhong Li and Derek Hoiem.
\newblock {Learning without Forgetting}.
\newblock \emph{IEEE Transactions on Pattern Analysis and Machine
  Intelligence}, 2017.

\bibitem[Lin(1992)]{experience_replay}
Long-Ji Lin.
\newblock Self-improving reactive agents based on reinforcement learning,
  planning and teaching.
\newblock \emph{Mach. Learn.}, 8\penalty0 (3–4):\penalty0 293–321, May
  1992.
\newblock ISSN 0885-6125.
\newblock \doi{10.1007/BF00992699}.

\bibitem[Lopez-Paz \& Ranzato(2017)Lopez-Paz and Ranzato]{Lopez-Paz}
David Lopez-Paz and Marc '~Aurelio Ranzato.
\newblock {Gradient Episodic Memory for Continual Learning}.
\newblock In \emph{Advances in Neural Information Processing Systems}, 2017.

\bibitem[Mankowitz et~al.(2018)Mankowitz, {\v{Z}}{\'\i}dek, Barreto, Horgan,
  Hessel, Quan, Oh, van Hasselt, Silver, and Schaul]{mankowitz2018unicorn}
Daniel~J Mankowitz, Augustin {\v{Z}}{\'\i}dek, Andr{\'e} Barreto, Dan Horgan,
  Matteo Hessel, John Quan, Junhyuk Oh, Hado van Hasselt, David Silver, and Tom
  Schaul.
\newblock Unicorn: Continual learning with a universal, off-policy agent.
\newblock \emph{arXiv preprint arXiv:1802.08294}, 2018.

\bibitem[Mendez et~al.(2020)Mendez, Wang, and Eaton]{Mendez2020}
Jorge~A Mendez, Boyu Wang, and Eric Eaton.
\newblock {Lifelong Policy Gradient Learning of Factored Policies for Faster
  Training Without Forgetting}.
\newblock In \emph{Advances in Neural Information Processing Systems}, 2020.

\bibitem[Mnih et~al.(2015)Mnih, Kavukcuoglu, Silver, Rusu, Veness, Bellemare,
  Graves, Riedmiller, Fidjeland, Ostrovski, Petersen, Beattie, Sadik,
  Antonoglou, King, Kumaran, Wierstra, Legg, and Hassabis]{Mnih}
Volodymyr Mnih, Koray Kavukcuoglu, David Silver, Andrei~A Rusu, Joel Veness,
  Marc~G Bellemare, Alex Graves, Martin Riedmiller, Andreas~K Fidjeland, Georg
  Ostrovski, Stig Petersen, Charles Beattie, Amir Sadik, Ioannis Antonoglou,
  Helen King, Dharshan Kumaran, Daan Wierstra, Shane Legg, and Demis Hassabis.
\newblock {Human-level control through deep reinforcement learning}.
\newblock \emph{Nature}, 2015.
\newblock \doi{10.1038/nature14236}.

\bibitem[Nagabandi et~al.(2018)Nagabandi, Finn, and Levine]{nagabandi2018deep}
Anusha Nagabandi, Chelsea Finn, and Sergey Levine.
\newblock Deep online learning via meta-learning: Continual adaptation for
  model-based rl.
\newblock \emph{arXiv preprint arXiv:1812.07671}, 2018.

\bibitem[Nguyen et~al.(2018)Nguyen, Li, Bui, and Turner]{vcl}
Cuong~V. Nguyen, Yingzhen Li, Thang~D. Bui, and Richard~E. Turner.
\newblock Variational continual learning.
\newblock In \emph{International Conference on Learning Representations}, 2018.

\bibitem[Nguyen et~al.(2021)Nguyen, Orbell, Lennon, Moon, Vigneau, Camenzind,
  Yu, Zumb{\"u}hl, Briggs, Osborne, et~al.]{nguyen2021deep}
V~Nguyen, SB~Orbell, Dominic~T Lennon, Hyungil Moon, Florian Vigneau, Leon~C
  Camenzind, Liuqi Yu, Dominik~M Zumb{\"u}hl, G~Andrew~D Briggs, Michael~A
  Osborne, et~al.
\newblock Deep reinforcement learning for efficient measurement of quantum
  devices.
\newblock \emph{npj Quantum Information}, 7\penalty0 (1):\penalty0 1--9, 2021.

\bibitem[OpenAI et~al.(2018)OpenAI, Andrychowicz, Baker, Chociej,
  J{\'{o}}zefowicz, McGrew, Pachocki, Pachocki, Petron, Plappert, Powell, Ray,
  Schneider, Sidor, Tobin, Welinder, Weng, and Zaremba]{dexterity}
OpenAI, Marcin Andrychowicz, Bowen Baker, Maciek Chociej, Rafal
  J{\'{o}}zefowicz, Bob McGrew, Jakub~W. Pachocki, Jakub Pachocki, Arthur
  Petron, Matthias Plappert, Glenn Powell, Alex Ray, Jonas Schneider, Szymon
  Sidor, Josh Tobin, Peter Welinder, Lilian Weng, and Wojciech Zaremba.
\newblock Learning dexterous in-hand manipulation.
\newblock \emph{CoRR}, abs/1808.00177, 2018.

\bibitem[Parker-Holder et~al.(2022)Parker-Holder, Jiang, Dennis, Samvelyan,
  Foerster, Grefenstette, and Rockt{\"a}schel]{parker2022evolving}
Jack Parker-Holder, Minqi Jiang, Michael Dennis, Mikayel Samvelyan, Jakob
  Foerster, Edward Grefenstette, and Tim Rockt{\"a}schel.
\newblock Evolving curricula with regret-based environment design.
\newblock \emph{arXiv preprint arXiv:2203.01302}, 2022.

\bibitem[Pathak et~al.(2017)Pathak, Agrawal, Efros, and
  Darrell]{pathak2017curiosity}
Deepak Pathak, Pulkit Agrawal, Alexei~A Efros, and Trevor Darrell.
\newblock Curiosity-driven exploration by self-supervised prediction.
\newblock In \emph{International conference on machine learning}, pp.\
  2778--2787. PMLR, 2017.

\bibitem[Pathak et~al.(2019)Pathak, Gandhi, and Gupta]{pathak2019self}
Deepak Pathak, Dhiraj Gandhi, and Abhinav Gupta.
\newblock Self-supervised exploration via disagreement.
\newblock In \emph{International conference on machine learning}, pp.\
  5062--5071. PMLR, 2019.

\bibitem[Powers et~al.(2021)Powers, Xing, Kolve, Mottaghi, and
  Gupta]{powers2021cora}
Sam Powers, Eliot Xing, Eric Kolve, Roozbeh Mottaghi, and Abhinav Gupta.
\newblock Cora: Benchmarks, baselines, and metrics as a platform for continual
  reinforcement learning agents.
\newblock \emph{arXiv preprint arXiv:2110.10067}, 2021.

\bibitem[Ring(1994)]{ring1994continual}
Mark~Bishop Ring.
\newblock \emph{Continual learning in reinforcement environments}.
\newblock PhD thesis, University of Texas at Austin, 1994.

\bibitem[Rolnick et~al.(2019)Rolnick, Ahuja, Schwarz, Lillicrap, and
  Wayne]{clear}
David Rolnick, Arun Ahuja, Jonathan Schwarz, Timothy Lillicrap, and Gregory
  Wayne.
\newblock Experience replay for continual learning.
\newblock In \emph{Advances in Neural Information Processing Systems 32}, pp.\
  350--360. 2019.

\bibitem[Rusu et~al.(2016)Rusu, Rabinowitz, Desjardins, Soyer, Kirkpatrick,
  Kavukcuoglu, Pascanu, and Hadsell]{progressivenets}
Andrei~A. Rusu, Neil~C. Rabinowitz, Guillaume Desjardins, Hubert Soyer, James
  Kirkpatrick, Koray Kavukcuoglu, Razvan Pascanu, and Raia Hadsell.
\newblock Progressive neural networks.
\newblock \emph{CoRR}, abs/1606.04671, 2016.

\bibitem[Samvelyan et~al.(2021)Samvelyan, Kirk, Kurin, Parker-Holder, Jiang,
  Hambro, Petroni, Kuttler, Grefenstette, and
  Rockt{\"a}schel]{samvelyan2021minihack}
Mikayel Samvelyan, Robert Kirk, Vitaly Kurin, Jack Parker-Holder, Minqi Jiang,
  Eric Hambro, Fabio Petroni, Heinrich Kuttler, Edward Grefenstette, and Tim
  Rockt{\"a}schel.
\newblock Minihack the planet: A sandbox for open-ended reinforcement learning
  research.
\newblock In \emph{Thirty-fifth Conference on Neural Information Processing
  Systems Datasets and Benchmarks Track (Round 1)}, 2021.
\newblock URL \url{https://openreview.net/forum?id=skFwlyefkWJ}.

\bibitem[Schaul et~al.(2015)Schaul, Horgan, Gregor, and
  Silver]{schaul2015universal}
Tom Schaul, Daniel Horgan, Karol Gregor, and David Silver.
\newblock Universal value function approximators.
\newblock In \emph{International conference on machine learning}, pp.\
  1312--1320. PMLR, 2015.

\bibitem[Schrittwieser et~al.(2019)Schrittwieser, Antonoglou, Hubert, Simonyan,
  Sifre, Schmitt, Guez, Lockhart, Hassabis, Graepel, Lillicrap, and
  Silver]{muzero}
Julian Schrittwieser, Ioannis Antonoglou, Thomas Hubert, Karen Simonyan,
  Laurent Sifre, Simon Schmitt, Arthur Guez, Edward Lockhart, Demis Hassabis,
  Thore Graepel, Timothy~P. Lillicrap, and David Silver.
\newblock Mastering atari, go, chess and shogi by planning with a learned
  model.
\newblock \emph{CoRR}, abs/1911.08265, 2019.

\bibitem[Schwarz et~al.(2018)Schwarz, Czarnecki, Luketina, Grabska{-}Barwinska,
  Teh, Pascanu, and Hadsell]{progress_compress}
Jonathan Schwarz, Wojciech Czarnecki, Jelena Luketina, Agnieszka
  Grabska{-}Barwinska, Yee~Whye Teh, Razvan Pascanu, and Raia Hadsell.
\newblock Progress {\&} compress: {A} scalable framework for continual
  learning.
\newblock In Jennifer~G. Dy and Andreas Krause (eds.), \emph{Proceedings of the
  35th International Conference on Machine Learning, {ICML} 2018,
  Stockholmsm{\"{a}}ssan, Stockholm, Sweden, July 10-15, 2018}, volume~80 of
  \emph{Proceedings of Machine Learning Research}, pp.\  4535--4544. {PMLR},
  2018.

\bibitem[Sekar et~al.(2020)Sekar, Rybkin, Daniilidis, Abbeel, Hafner, and
  Pathak]{sekar2020planning}
Ramanan Sekar, Oleh Rybkin, Kostas Daniilidis, Pieter Abbeel, Danijar Hafner,
  and Deepak Pathak.
\newblock Planning to explore via self-supervised world models.
\newblock In \emph{International Conference on Machine Learning}, pp.\
  8583--8592. PMLR, 2020.

\bibitem[Shi et~al.(2015)Shi, Chen, Wang, Yeung, Wong, and
  Woo]{shi2015convolutional}
Xingjian Shi, Zhourong Chen, Hao Wang, Dit-Yan Yeung, Wai-Kin Wong, and
  Wang-chun Woo.
\newblock Convolutional lstm network: A machine learning approach for
  precipitation nowcasting.
\newblock \emph{Advances in neural information processing systems}, 28, 2015.

\bibitem[Shin et~al.(2017)Shin, Lee, Kim, and Kim]{Shin2017}
Hanul Shin, Jung~Kwon Lee, Jaehong Kim, and Jiwon Kim.
\newblock {Continual Learning with Deep Generative Replay}.
\newblock In \emph{Advances in Neural Information Processing Systems}, 2017.

\bibitem[Steinparz et~al.(2022)Steinparz, Schmied, Paischer, Dinu, Patil,
  Bitto-Nemling, Eghbal-zadeh, and Hochreiter]{steinparz2022reactive}
Christian Steinparz, Thomas Schmied, Fabian Paischer, Marius-Constantin Dinu,
  Vihang Patil, Angela Bitto-Nemling, Hamid Eghbal-zadeh, and Sepp Hochreiter.
\newblock Reactive exploration to cope with non-stationarity in lifelong
  reinforcement learning.
\newblock \emph{arXiv preprint arXiv:2207.05742}, 2022.

\bibitem[Sutton(1991)]{sutton1991dyna}
Richard~S Sutton.
\newblock Dyna, an integrated architecture for learning, planning, and
  reacting.
\newblock \emph{ACM Sigart Bulletin}, 2\penalty0 (4):\penalty0 160--163, 1991.

\bibitem[Sutton \& Barto(2018)Sutton and Barto]{sutton2018reinforcement}
Richard~S Sutton and Andrew~G Barto.
\newblock \emph{Reinforcement learning: An introduction}.
\newblock MIT press, 2018.

\bibitem[Team et~al.(2021)Team, Stooke, Mahajan, Barros, Deck, Bauer,
  Sygnowski, Trebacz, Jaderberg, Mathieu, et~al.]{team2021open}
Open Ended~Learning Team, Adam Stooke, Anuj Mahajan, Catarina Barros, Charlie
  Deck, Jakob Bauer, Jakub Sygnowski, Maja Trebacz, Max Jaderberg, Michael
  Mathieu, et~al.
\newblock Open-ended learning leads to generally capable agents.
\newblock \emph{arXiv preprint arXiv:2107.12808}, 2021.

\bibitem[Thrun \& Mitchell(1995)Thrun and Mitchell]{THRUN199525}
Sebastian Thrun and Tom~M. Mitchell.
\newblock Lifelong robot learning.
\newblock \emph{Robotics and Autonomous Systems}, 15\penalty0 (1):\penalty0
  25--46, 1995.
\newblock ISSN 0921-8890.
\newblock \doi{https://doi.org/10.1016/0921-8890(95)00004-Y}.
\newblock URL
  \url{https://www.sciencedirect.com/science/article/pii/092188909500004Y}.
\newblock The Biology and Technology of Intelligent Autonomous Agents.

\bibitem[Van~de Ven \& Tolias(2019)Van~de Ven and Tolias]{van2019three}
Gido~M Van~de Ven and Andreas~S Tolias.
\newblock Three scenarios for continual learning.
\newblock \emph{arXiv preprint arXiv:1904.07734}, 2019.

\bibitem[Van Den~Oord et~al.(2017)Van Den~Oord, Vinyals, et~al.]{van2017neural}
Aaron Van Den~Oord, Oriol Vinyals, et~al.
\newblock Neural discrete representation learning.
\newblock \emph{Advances in neural information processing systems}, 30, 2017.

\bibitem[Van~Seijen et~al.(2020)Van~Seijen, Nekoei, Racah, and
  Chandar]{van2020loca}
Harm Van~Seijen, Hadi Nekoei, Evan Racah, and Sarath Chandar.
\newblock The loca regret: a consistent metric to evaluate model-based behavior
  in reinforcement learning.
\newblock \emph{Advances in Neural Information Processing Systems},
  33:\penalty0 6562--6572, 2020.

\bibitem[Vitter(1985)]{Vitter1985RandomSW}
Jeffrey~Scott Vitter.
\newblock Random sampling with a reservoir.
\newblock \emph{ACM Trans. Math. Softw.}, 11:\penalty0 37--57, 1985.

\bibitem[Wan et~al.(2022)Wan, Rahimi-Kalahroudi, Rajendran, Momennejad,
  Chandar, and Van~Seijen]{wan2022towards}
Yi~Wan, Ali Rahimi-Kalahroudi, Janarthanan Rajendran, Ida Momennejad, Sarath
  Chandar, and Harm~H Van~Seijen.
\newblock Towards evaluating adaptivity of model-based reinforcement learning
  methods.
\newblock In \emph{International Conference on Machine Learning}, pp.\
  22536--22561. PMLR, 2022.

\bibitem[Wang et~al.(2019)Wang, Lehman, Clune, and Stanley]{wang2019paired}
Rui Wang, Joel Lehman, Jeff Clune, and Kenneth~O Stanley.
\newblock Paired open-ended trailblazer (poet): Endlessly generating
  increasingly complex and diverse learning environments and their solutions.
\newblock \emph{arXiv preprint arXiv:1901.01753}, 2019.

\bibitem[Wolczyk et~al.(2021)Wolczyk, Zajac, Pascanu, Kucinski, and
  Milos]{wolczyk2021continual}
Maciej Wolczyk, Michal Zajac, Razvan Pascanu, Lukasz Kucinski, and Piotr Milos.
\newblock Continual world: {A} robotic benchmark for continual reinforcement
  learning.
\newblock In Marc'Aurelio Ranzato, Alina Beygelzimer, Yann~N. Dauphin, Percy
  Liang, and Jennifer~Wortman Vaughan (eds.), \emph{Advances in Neural
  Information Processing Systems 34: Annual Conference on Neural Information
  Processing Systems 2021, NeurIPS 2021, December 6-14, 2021, virtual}, pp.\
  28496--28510, 2021.
\newblock URL
  \url{https://proceedings.neurips.cc/paper/2021/hash/ef8446f35513a8d6aa2308357a268a7e-Abstract.html}.

\bibitem[Xie et~al.(2020)Xie, Harrison, and Finn]{xie2020deep}
Annie Xie, James Harrison, and Chelsea Finn.
\newblock Deep reinforcement learning amidst lifelong non-stationarity.
\newblock \emph{arXiv preprint arXiv:2006.10701}, 2020.

\bibitem[Zenke et~al.(2017)Zenke, Poole, and Ganguli]{Zenke2017}
Friedemann Zenke, Ben Poole, and Surya Ganguli.
\newblock {Continual Learning Through Synaptic Intelligence}.
\newblock In \emph{International Conference on Machine Learning}, 2017.

\end{thebibliography}

\bibliographystyle{collas2023_conference}

\appendix

\clearpage

\begin{appendices}

\crefalias{section}{appsec}
\crefalias{subsection}{appsec}
\crefalias{subsubsection}{appsec}

\setcounter{equation}{0}
\renewcommand{\theequation}{\thesection.\arabic{equation}}

\onecolumn

\section*{\LARGE Appendix}
\label{sec:appendix}

\section*{Table of Contents}
\vspace*{-10pt}
\startcontents[sections]
\printcontents[sections]{l}{1}{\setcounter{tocdepth}{2}}

\section{Continual Supervised Learning}

\subsection{Definition}
\label{app:cl_def}
Continual Learning (CL) is a setting whereby a model must master a set of tasks sequentially while maintaining performance across all previously learned tasks. Other important objectives are to develop scalable CL methods which are able to transfer knowledge from previous tasks to aid the learning of new tasks, known as forward transfer. Traditionally, in supervised CL, the model is sequentially shown $T$ tasks, denoted $\mathcal{T}_{\tau}$ for $\tau= 1, \ldots, T$. Each task, $\mathcal{T}_{\tau}$, is comprised of a dataset $\mathcal{D}_{\tau} = \left \{ (\vx_{i}, y_{i}) \right  \}_{i = 1}^{ N_{\tau}}$ which a neural network is required to learn from. More generally, a task is denoted by a tuple comprised of the conditional and marginal distributions, $\{ p_{\tau}(y|\rvx), p_{\tau}(\rvx)\}$. After task $\tau$, the model will lose access to the training dataset for $\mathcal{T}_{\tau}$, however, its performance will be continually evaluated on all tasks $\mathcal{T}_i$ for $i \leq \tau$. For a comprehensive review of CL scenarios see~\citep{hsu2018re,van2019three}.

\subsection{Related Works}
\label{app:cl_related_works}
Here, we briefly describe CL methods. One approach to CL referred to as \emph{regularization approaches} regularizes a NN's weights to ensure that optimizing for a new task finds a solution that is ``close'' to the previous task's \citep{ewc, vcl, Zenke2017}. Working with functions can be easier than with NN weights and so task functions can be regularized to ensure that learning new function mappings are ``close'' across tasks \citep{Li2017, Benjamin2019, buzzega2020dark}. By contrast, \emph{expansion approaches} add new NN components to enable learning new tasks while preserving components for specific tasks \citep{progressivenets, lee2020neural}. \emph{Memory approaches} replay data from previous tasks when learning the current task. This can be performed with a generative model \citep{Shin2017}. Or samples from previous tasks (\emph{memories}) \citep{Lopez-Paz, Aljundi2019, ChaudhryTinyEps}.

\section{Continual Reinforcement Learning Metrics}
\label{app:metrics}
We describe in detail how to calculate the continual reinforcement learning metrics used extensively throughout this manuscript.

\subsection{Average Performance} 
This measures how well a CRL method performs on all tasks at the end of the task sequence. The task performance is $p_{\tau}(t) = [-1, 1]$ for all $\tau < T$. Since we have a reward of $+1$ for completing the task and $-1$ for being killed by a monster or falling into lava. If each task is seen for $N$ environment steps and we have $T$ tasks and the $\tau$-th task is seen over the interval of steps $[(\tau-1) \times N, \tau \times N]$. The average final performance metric for our continual learning agent is defined as:
\begin{align}
    p(t_{f}) = \frac{1}{T} \sum_{\tau=1}^{T}p_{\tau}(t_f),
\end{align}
where $t_f= N \times T$ is the final timestep. %The average integrated performance averages the performance over the entire lifetime that the agent has seen a particular task:
%\begin{align}
%    p = \frac{1}{T} \sum_{\tau=1}^{T} \frac{1}{N} \int_{\tau-1 \times N}^{t_f}p_{\tau}(t) dt.
%\end{align}

\subsection{Forgetting}
The average forgetting is the performance difference after  interacting with a task versus the performance at the end of the final task. The average forgetting across all tasks is defined as:
\begin{align}
    F = \frac{1}{T}\sum_{\tau=1}^{T} F_{\tau} \quad \textrm{where} \quad F_{\tau} = p_{\tau}(\tau \times N) -  p_{\tau}(t_{f}).
\end{align}
The forgetting of the final $T$-th task is $F_{T} = 0$. If a CRL agent has better performance at the end of the task sequence compared to after $\tau$-th task at time-step $\tau \times N$ then $F_{\tau} < 0$. Note that both the average performance and the forgetting metrics are functions of $p_{\tau}(t_{f})$ so we expect anti-correlation between these two metrics.

\subsection{Forward Transfer} 
The forward transfer is the difference in task performance during continual learning compared to the single task performance. The forward transfer is defined:
\begin{align}
\label{eq:fwd_trans}
    FT &= \frac{1}{T} \sum_{\tau=1}^{T} FT_{\tau}
    \quad \textrm{where} \quad FT_{\tau} = \frac{ \textrm{AUC}_{\tau} - \textrm{AUC}_{\textrm{ref}_{\tau}}}{1 - \textrm{AUC}_{\tau}},
\end{align}
where $\textrm{AUC}$ denotes the area under the curve and is defined as:
\begin{align}
    \textrm{AUC}_{\tau} &= \frac{1}{N} \int^{\tau\times N}_{(\tau-1) \times N} p_{\tau}(t)dt 
     \quad \textrm{and} \quad  \textrm{AUC}_{\textrm{ref}_{\tau}} = \frac{1}{N}\int_{0}^{N} p_{\textrm{ref}_{\tau}}(t)dt.
\end{align}
$FT_{\tau} > 0$ means that the CRL agent achieves better performance on task $\tau$ during continual learning versus in isolation. So this metric measures how well a CRL agent transfers knowledge from previous tasks when learning a new task.

\section{Single Task experiments}

\begin{figure*}
    \centering
    \includegraphics[width=1.0\textwidth]{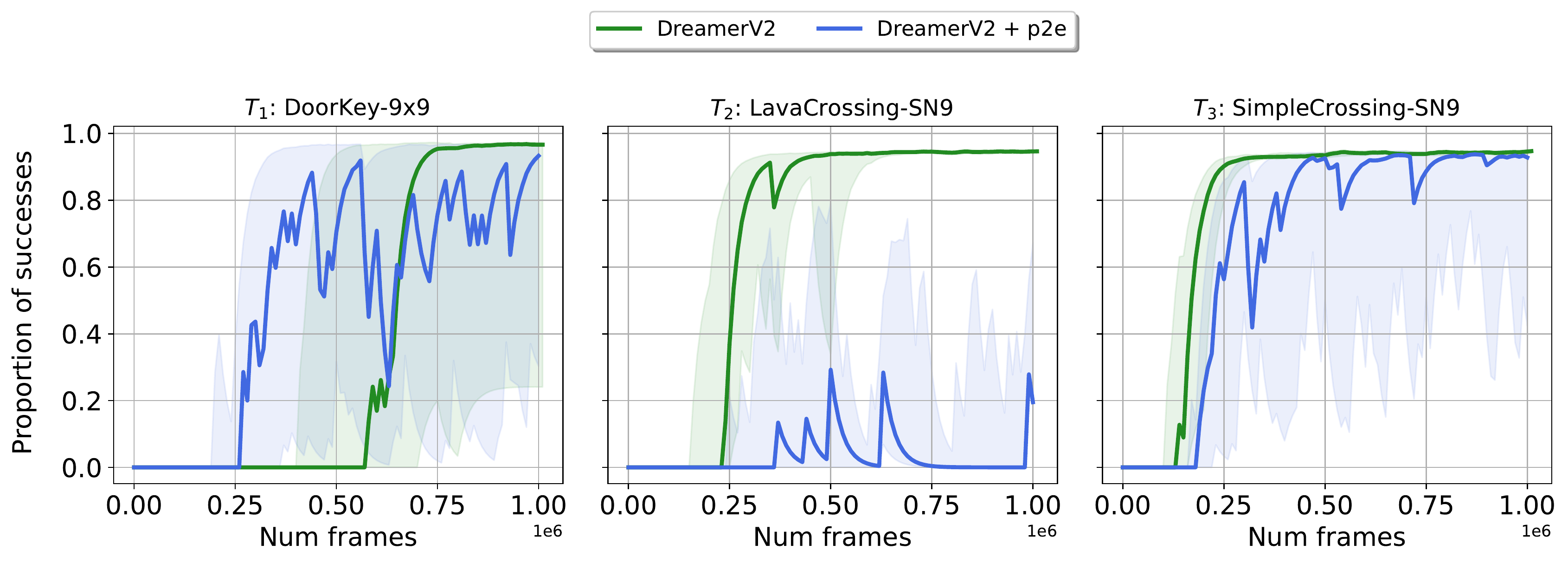}
    \caption{Single task performance of on individual tasks from the Minigrid CRL benchmark. All curves are a median and inter-quartile range over $20$ seeds.}
    \label{fig:dv2_minigrid_single}
\end{figure*}

\begin{figure*}
    \centering
    %\hspace{-5.0cm}
    \includegraphics[width=1.0\textwidth]{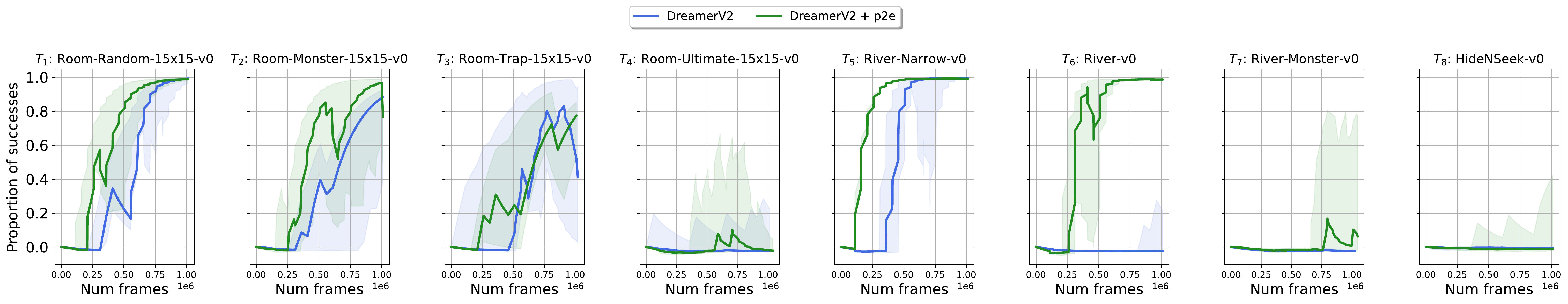}
    \caption{Single task performance on individual tasks from the Minihack CRL benchmark. All curves are a median and inter-quartile range over $10$ seeds.}
    \label{fig:dv2_minihack_single}
\end{figure*}

To assess the forward transfer of DreamerV2 for CRL we need the performance of each task as a reference~\cref{eq:fwd_trans}. Single task learning curves for Minigrid are shown in~\cref{fig:dv2_minigrid_single} and single task learning curves for all Minihack tasks are shown in~\cref{fig:dv2_minihack_single}.

\section{Further Experiments}

We introduce further experiments which are referenced in the main paper. In~\Cref{sec:app:4task_mh} we show learning curves for each individual task for $4$ task Minihack. In~\Cref{sec:app:task-aware} we show the results of the regularization based task-aware world model baseline. In~\Cref{sec:app:8task_mh} we show learning curves for each individual task from $8$ task Minihack experimental setup.  In~\Cref{sec:minihack_ablations} we explore various design choices required for DreamerV2 with Plan2Explore to get the best performance for CRL. In~\Cref{sec:minihack_buffer_sz} we explore how increasing the size of the experience replay buffer size affects performance in the Minihack CRL benchmark. These experiments are on $8$ tasks of Minihack. The $8$ tasks are a subset of those introduced in the CORA Minihack suite~\citep{powers2021cora} and are a superset of the $4$ tasks in the main paper. The $8$ tasks, in order, are: \RoomRandom{}, \RoomMonster{}, \RoomTrap{}, \RoomUltimate{}, \RiverNarrow{}, \River{}, \RiverMonster{}, and \HideNSeek{}.

\subsection{$4$ task Minihack}
\label{sec:app:4task_mh}
\begin{figure}
    \centering
    \includegraphics[width=0.95\textwidth]{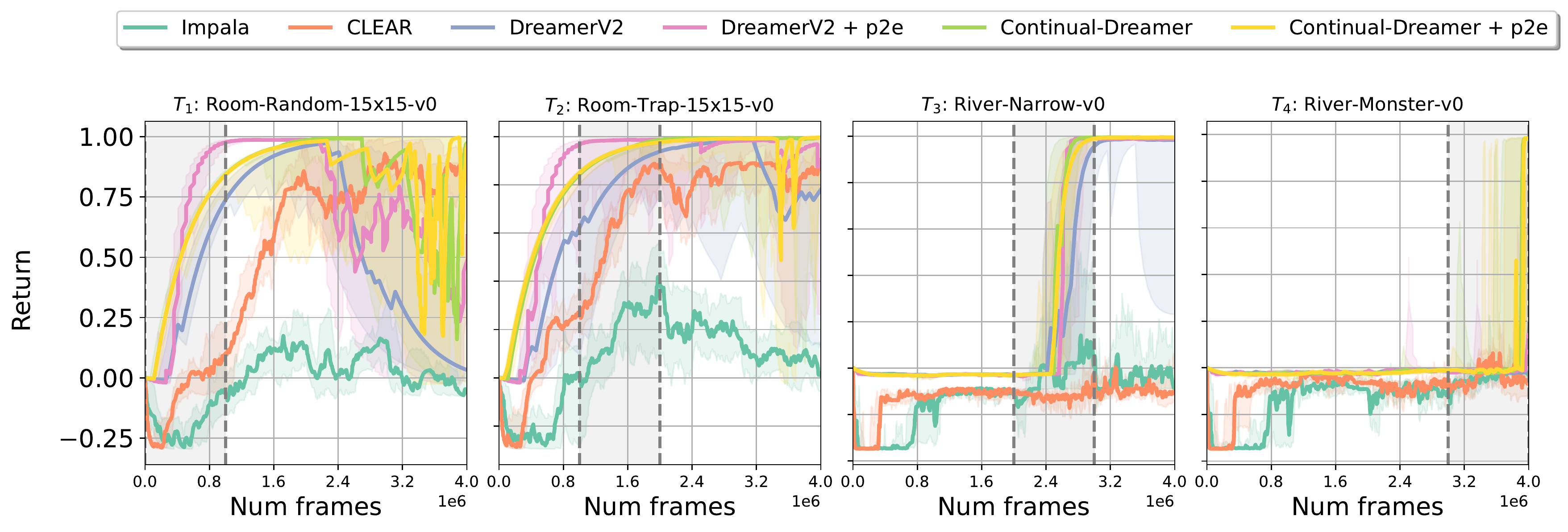}
    \caption{Detailed, per task, comparison of baselines and DreamerV2 variants that are presented in \cref{fig:minihack_cl} with average return. All curves are a median and inter-quartile range over $20$ seeds.}
    \label{fig:app:mh_4_task_return}
\end{figure}

\cref{fig:app:mh_4_task_return} shows the success rates of the baseline methods and DreamerV2 variants for each of the four tasks in the Minihack experiment~\cref{sec:results_minihack}. Continual Dreamer successfully balances retaining knowledge from previous tasks and learning new ones. On the other hand, the CLEAR baseline can only learn the first two tasks with a significant delay, while Impala struggles on all tasks. DreamerV2 and DreamerV2 + Plan2Explore perform poorly on the last task and exhibit more forgetting than Continual-Dreamer.

\subsection{Task-aware world-model baseline}
\label{sec:app:task-aware}
\begin{figure}
    \centering
    \includegraphics[width=0.6\textwidth]{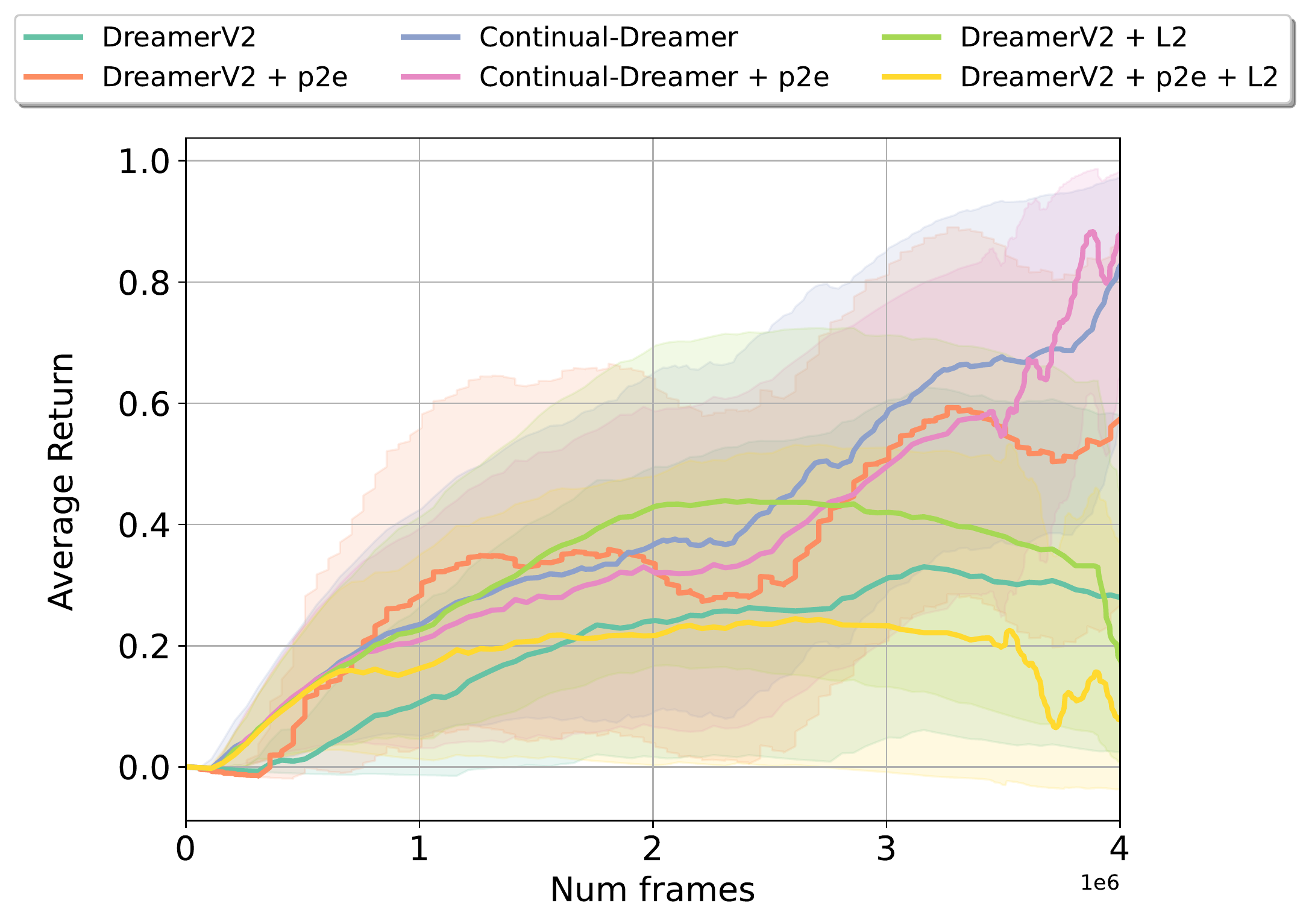}
    \caption{Comparison of the average return over all tasks between task-aware and task-agnostic approaches based on DreamerV2, on 4 Minihack tasks. All curves are IQM from \texttt{rliable} package across 10 seeds and 1000 bootstrap samples.}
    \label{fig:app:task_aware}
\end{figure}

Differences for task-agnostic and task-aware variants of DreamerV2 are shown in~\cref{fig:app:task_aware}. The task-aware variant based on $\normltwo$ regularization, underperforms in comparison to task-agnostic methods. The plausible explanation is that $\normltwo$ regularization makes the method too rigid to efficiently learn the two last tasks because of the excessive influence of the first tasks. We optimize the size of the $\normltwo$ scaling from the set $\{10^{-4}, 10^{-3}, 10^{-2}, 10^{-1}, 1, 10, 100\}$.

\subsection{Scaling to $8$ tasks}
\label{sec:app:8task_mh}

\begin{figure}
    \centering
    \includegraphics[width=0.95\textwidth]{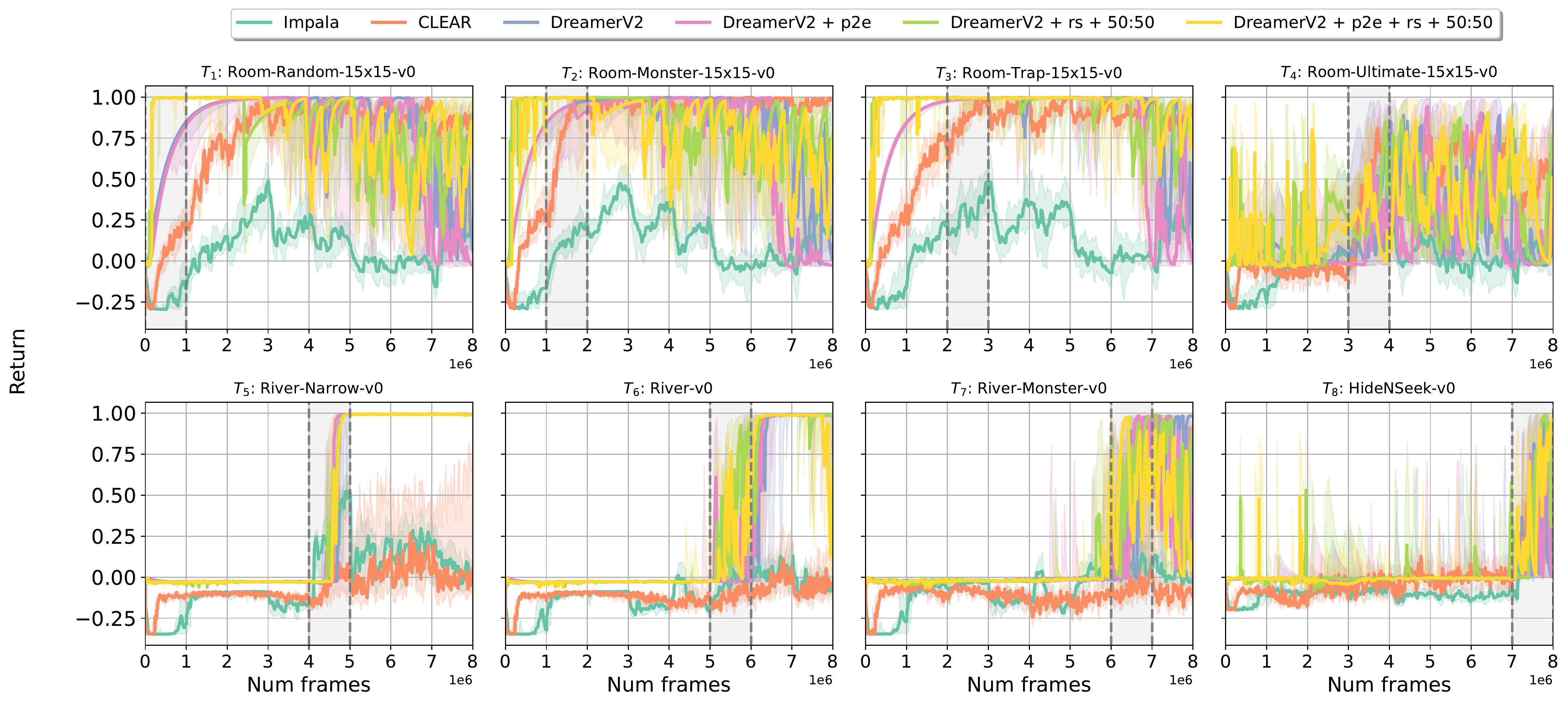}
    \caption{Learning curves for $8$ Minihack tasks for DreamerV2 and variants and Impala and CLEAR baselines. All curves are a median and interquartile range of $10$ seeds.}
    \label{fig:app:8task_mh}
\end{figure}

The results for 8 Minihack tasks are shown in~\cref{fig:app:8task_mh}. DreamerV2 variants display strong knowledge retention and effective learning on almost every task compared to baseline methods. CLEAR method struggle with the last 4 tasks, whereas Impala's performance is poor on every task. DreamerV2 and variants displays forgetting of the initial tasks, for which CLEAR retains the highest performance. However, CLEAR, in contrast to DreamerV2 variants, struggles to learn novel tasks. 

\subsection{DreamerV2 Ablation Experiments}
\label{sec:minihack_ablations}
We explore various design choices which come from the implementations of DreamerV2 \citep{hafner2020mastering} and Plan2Explore \citep{sekar2020planning}.
\begin{enumerate}
    \item The use of Plan2Explore as an intrinsic reward.
    \item World model learning by reconstructing the observations $\hat{o}_t$ only and not the observations, rewards, and discounts altogether.
    \item The use of the exploration policy to evaluate the performance on all current and past tasks rather than having a separate exploration and evaluation policy.
\end{enumerate}
The results are shown in~\cref{tab:minihack_ablation}. We decided to pick the model in the final line in~\cref{tab:minihack_ablation} to report the results in the main paper as they produce good results on Minihack with a relatively small standard deviation.

\begin{table}[]
    \centering
    \resizebox{\textwidth}{!}{\begin{tabular}{cccccc}
        \toprule
         Plan2Explore & $\hat{o}$ reconstruction only & $\pi_{exp} = \pi_{eval}$ & Avg. Performance ($\uparrow$) & Avg. Forgetting ($\downarrow$) & Avg. Forward Transfer ($\uparrow$) \\
         \midrule
         - & - & - & $0.09 \pm 0.07$ & $0.37 \pm 0.07$ & $0.56 \pm 0.86$ \\ 
         \ding{52} & - & - & $ 0.28 \pm 0.13$ & $0.13 \pm 0.08$ & $0.11 \pm 0.15$ \\
         \ding{52} & \ding{52} & - & $0.39 \pm 0.13$ & $0.19 \pm 0.16$ & $0.87 \pm 0.95$ \\ 
         \ding{52} & \ding{52} & \ding{52} & $0.38 \pm 0.03$ & $0.22 \pm 0.05$ & $0.76 \pm 0.25$ \\
         \bottomrule
    \end{tabular}}
    \caption{CRL metrics for different design decisions on DreamerV2 for the Minihack CRL benchmark of $8$ tasks. All metrics are an average and standard deviation over $5$ seeds. $\uparrow$ indicates better performance with higher numbers, and $\downarrow$ the opposite.}
    \label{tab:minihack_ablation}
\end{table}

\subsection{Stability versus Plasticity: Increasing the Size of the Replay Buffer}
\label{sec:minihack_buffer_sz}

\begin{figure*}
    \centering
    %\hspace{-5.0cm}
    %\includegraphics[width=0.95\textwidth]{plots/minihack_cl_metrics_vs_buffer_sz.pdf}
    \includegraphics[width=0.95\textwidth]{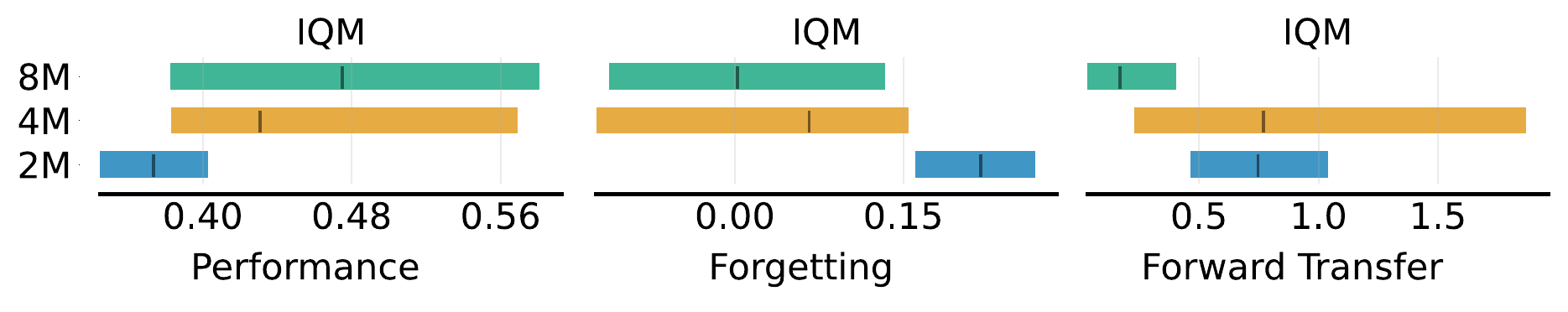}
    \caption{CRL metric package for DreamerV2 + Plan2Explore for the Minihack benchmark of $8$ tasks versus the experience replay buffer size of the world model for DreamerV2 + Plan2Explore. All metrics are an interquartile mean (IQM) over $5$ seeds with $1000$ bootstrap samples from the \texttt{rliable}.}
    \label{fig:dv2_buffer_sz}
\end{figure*}

By increasing the replay buffer size for world model learning for DreamerV2 + Plan2Explore we see that forgetting and average performance increase. However, the forward transfer simultaneously decreases,~\cref{fig:dv2_buffer_sz}. Additionally, by inspecting the learning curves we notice that the harder exploration tasks are not learned as the replay buffer size increases. This is an instance of the stability-plasticity trade-off in continual learning the larger buffer size enables better remembering but simultaneously prevents quickly learning the new tasks. 

\subsection{Interference}
\label{sec:4R_interfernce}
\begin{wrapfigure}{r}{0.45\textwidth}
    \centering
    \vspace{-0.8cm}
    \includegraphics[width=0.45\textwidth]{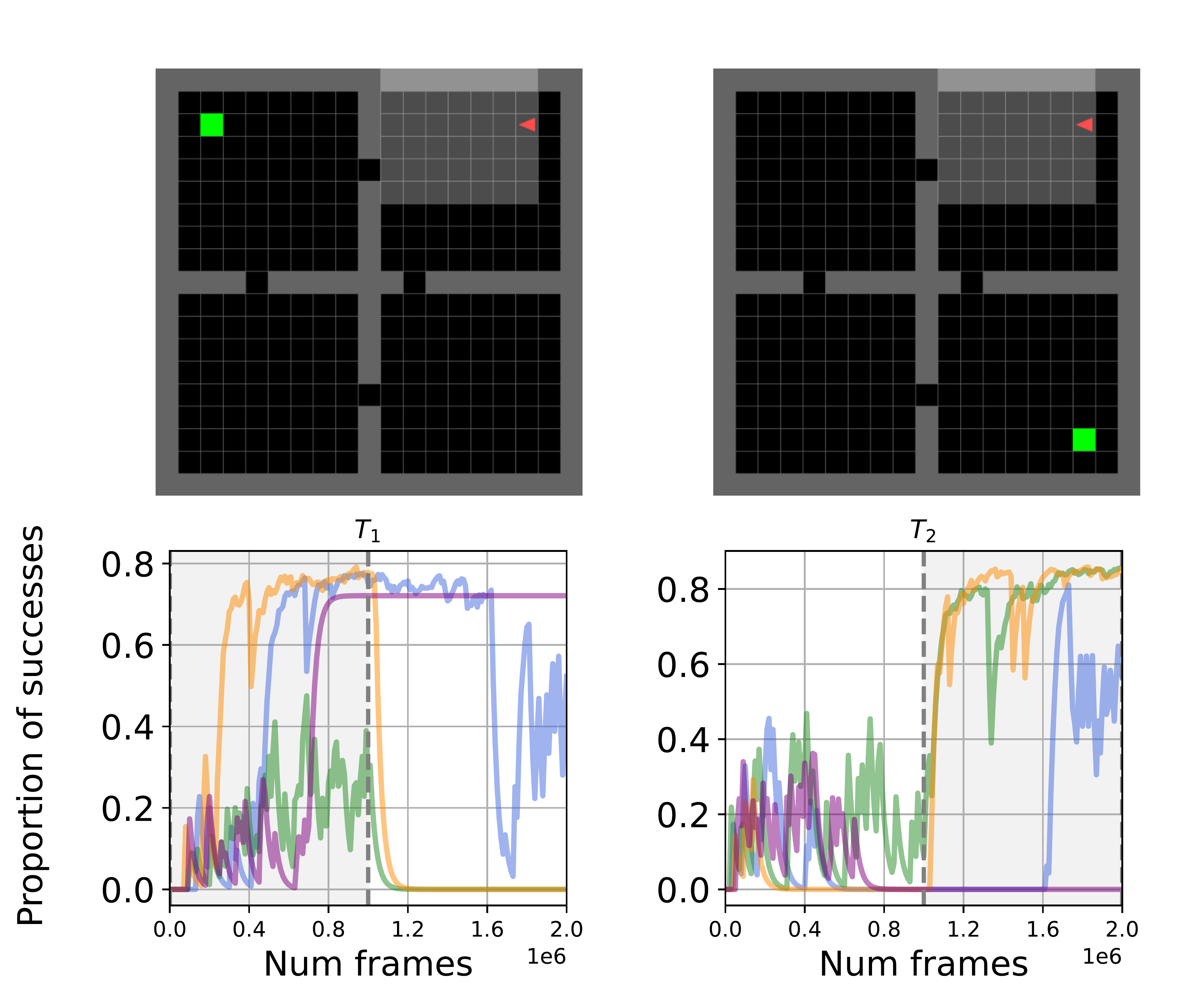}
    \caption{\textbf{Top}, $2$ \FourRooms{} environment on with fixed agent start location, agent start orientation, and obstacles. Only the goal or reward function changes from one task to the next. \textbf{Bottom}, $4$ separate success rate learning curves for different random seeds in different colours for DreamerV2 + p2e.}
    \vspace{-0.5cm}
    \label{fig:interference}
\end{wrapfigure}

We observe interference explicitly using DreamerV2 with Plan2Explore on the Minigrid \FourRooms{} environment where we change the reward function or goal location from one task to another~\cref{fig:interference}. We train for $1$M steps on the first task, then $1$M steps on the second task where the goal location has changed with all else the same. From the learning curves for each individual run, we can see how only one task will ever be solved with simultaneously poor performance in the second task (with the exception of one seed, in blue, which switches between solving and not solving each task).

%\piotrm{I would probably make this paragraph shorter, stating that 'we do not tackle interference and defer it to a further study' and put the example to the appendix. But this is probably minor.} \sam{all reviewers were pretty keen we discuss this, also given we put this in the rebuttal, maybe the reviewers expect this in the revise and resubmit.}

\subsection{Decreasing the Replay Buffer Size}
\label{sec:minigrid_small_replay_buffer}

We decrease the size of the replay buffer for DreamerV2 and variants to see how well it is able to manage environment reconstruction in the face of continual learning and decreasing replay buffer size. We consider the $3$ task Minigrid continual learning problem and decrease the size of the replay buffer from $2 \times 10^{6}$ transitions which is used in~\cref{sec:results_minigrid} to replay buffers of size $\{10^{4}, 10^{5}, 10^{6} \}$ transitions.

\begin{figure}
\centering
\begin{subfigure}
  \centering
    \includegraphics[width=0.9\textwidth]{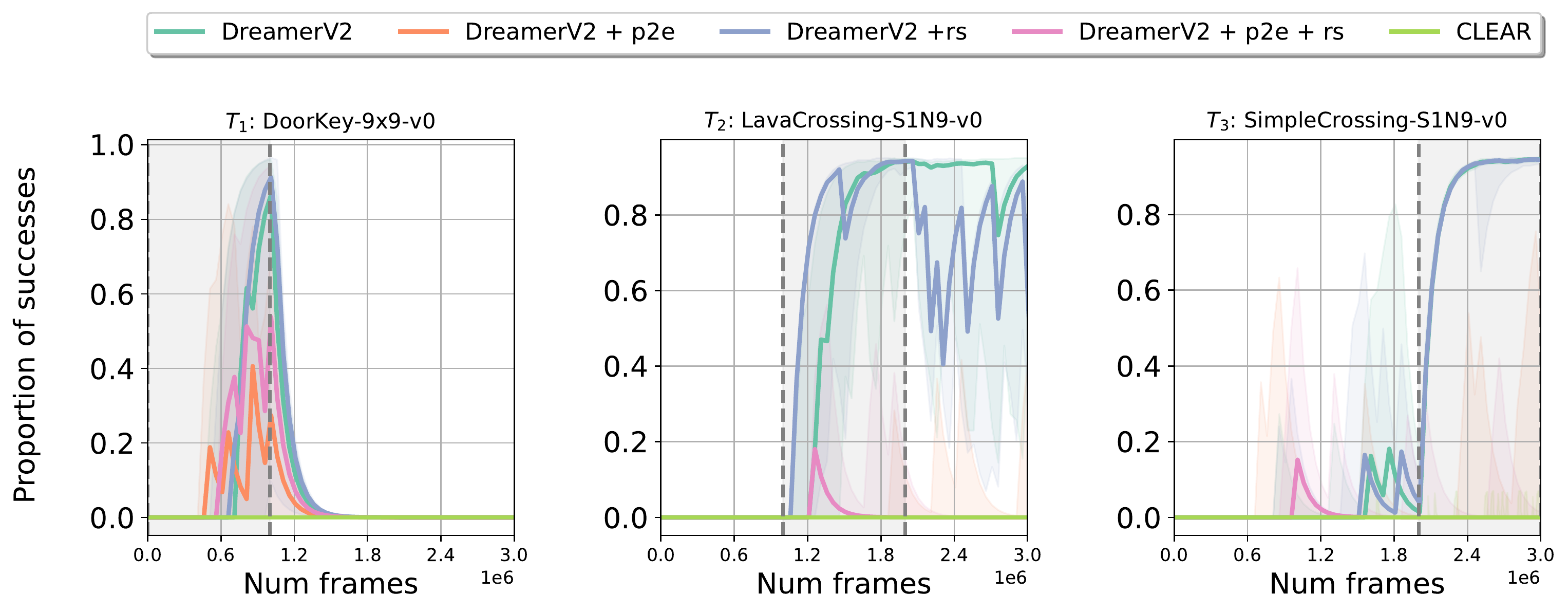}
    \caption*{Replay buffer size $10^4$.}
    \label{fig:dv2_buffer_sz_tiny}
\end{subfigure}

\begin{subfigure}
  \centering
    \includegraphics[width=0.9\textwidth]{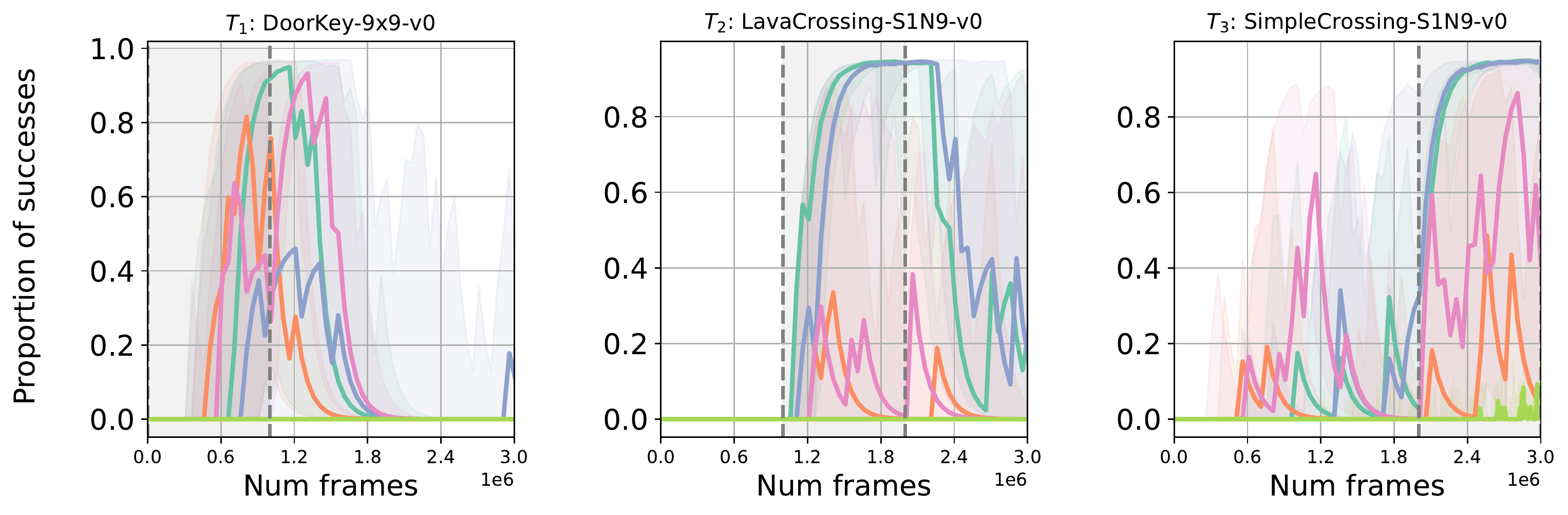}
    \caption*{Replay buffer size $10^5$.}
    \label{fig:dv2_buffer_sz_small}
\end{subfigure}

\begin{subfigure}
  \centering
    \includegraphics[width=0.9\textwidth]{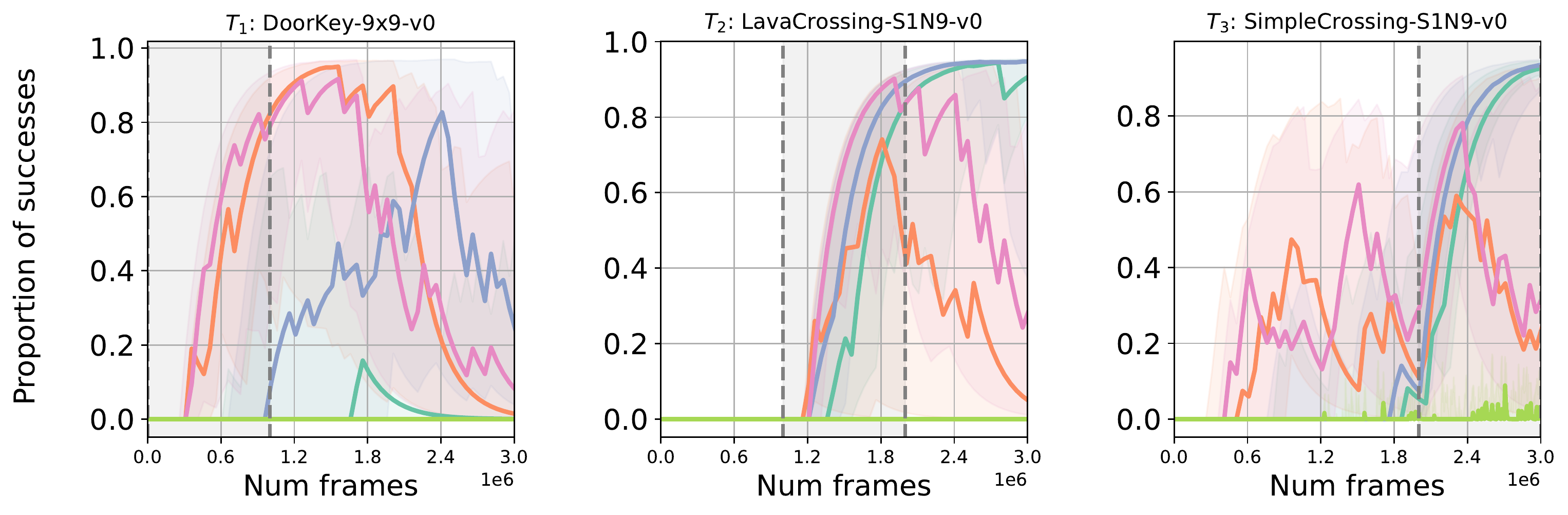}
    \caption*{Replay buffer size $10^6$.}
    \label{fig:dv2_buffer_sz_normal}
\end{subfigure}

\caption{Learning curves for continual learning on $3$ different Minigrid tasks with $1$M environment interactions per task before changing task. The replay buffer is decreased from the experiments in~\cref{sec:results_minigrid} to $10^4$ transitions in the top row, $10^5$ in the middle row, and $10^6$ in the bottom row. All runs are medians and interquartile ranges of over $10$ different runs with different seeds.}
\label{fig:mg_replay_buffer_sizes}
\end{figure}

From the results in~\cref{fig:mg_replay_buffer_sizes}, we can see that DreamerV2 and its variants under-perform with small replay buffers of size $10^{4}$ and $10^{5}$. DreamerV2 with Plan2Explore is unable to learn with such small replay buffers. DreamerV2 by itself is better at learning under small replay buffers. DreamerV2 with Plan2Explore learns to solve the difficult \DoorKey{} problem only when it has a replay buffer size of $10^6$. We can also see that reservoir sampling helps against catastrophic forgetting for both Dreamer and DreamerV2 with Plan2Explore with replay buffer sizes of $10^5$ and $10^6$.
%\piotrm{The analysis would be more meaningful, if we compare to the single-task baseline. We should have a clear message, if the failure is due to the fact that even single-task does not work, or there is some CL issue.} \sam{I agree. But, we included this analysis in the rebuttal with only CL experiments. The meta-reviewer seemed to appreciate it, so we should keep it in?}

\subsection{Analysis of Replay Buffer Sampling Methods}
\label{sec:sampling_distribution}
\begin{figure}
    \centering
    \includegraphics[width=0.8\textwidth]{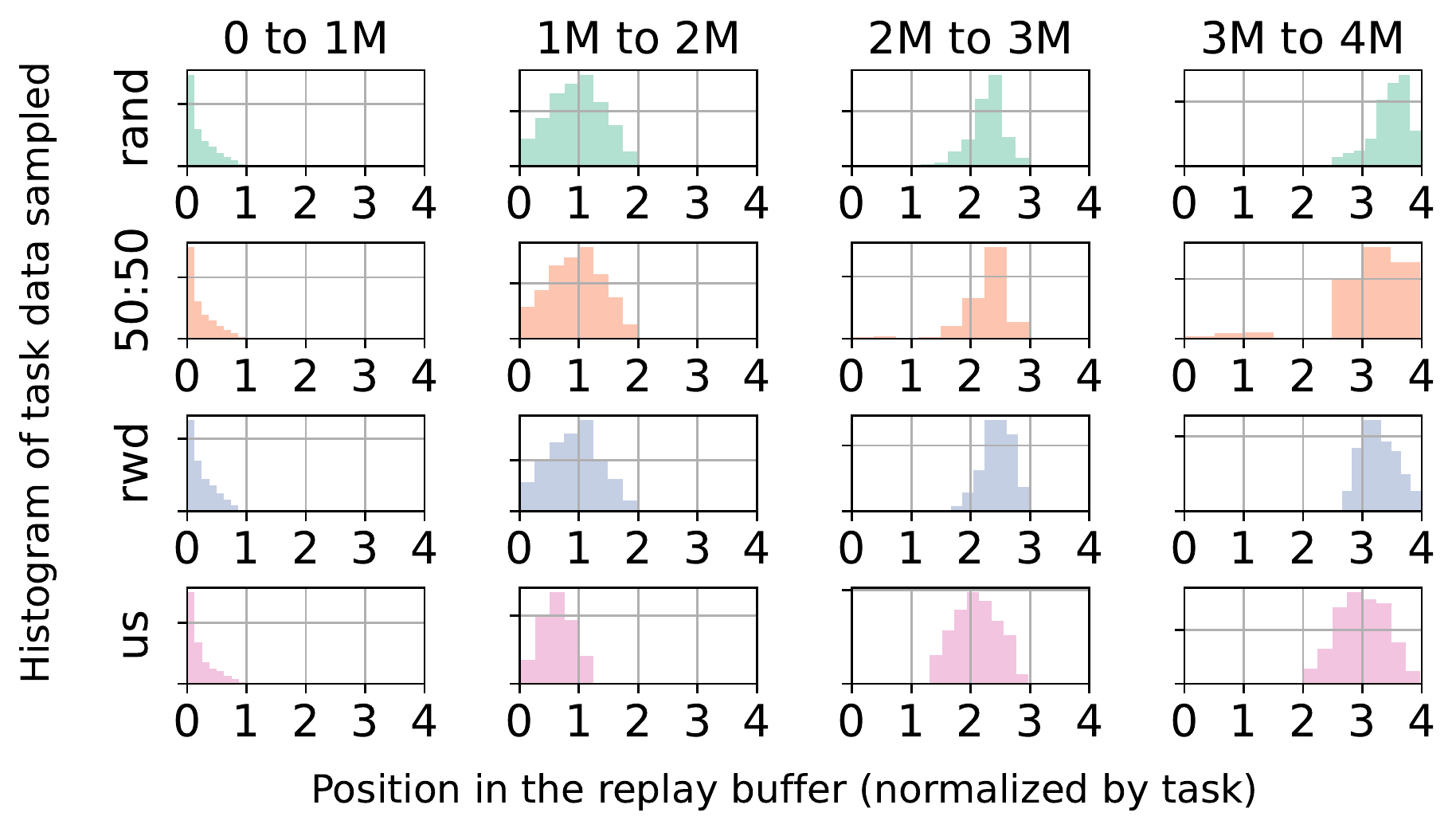}
    \caption{Distribution of episode indexes which are sampled to construct a minibatch for various sampling methods (rows) while the world-model is interacting with each task (columns).}
    \label{fig:sampling_distribution}
\end{figure}

We analyze the workings of the different minibatch construction sampling methods: random sampling, \emph{us}, \emph{rwd} and \emph{50:50} sampling. For $50:50$ sampling we also employ reservoir sampling (since in our experiments in~\cref{sec:experiments} we employ it to add plasticity to \emph{rs}). All other sampling methods use a FIFO replay buffer. We plot histograms of the episode index which is sampled for mini-batch construction for world-model learning and initiating the Actor-Critic learning. We normalized the replay buffer episode indexes, which are sampled so that they lie in the range $[0, 4]$ to indicate which task the episode corresponds to. So all the episode indexes which were sampled while the agent interacted with a particular task are visualized in a distinct histogram (columns) for various sampling methods (rows)~\cref{fig:sampling_distribution}.

We can see from~\cref{fig:sampling_distribution} that the sampling from the replay buffer is similar across all sampling methods for the first task. The distributions for random sampling are not uniform since only episodes which are of length $>50$ are stored in the replay buffer, so the histograms' first row is not flat since certain indexes will not be sampled as a result. The distribution for \emph{rwd} and \emph{us} is similar to random sampling for the first task. This must mean that the uncertainty from Plan2Explore is quite uniform for the first task, perhaps even the uncertainty is larger for the earlier episodes in the replay buffer for the first task (row 4, column 1~\cref{fig:sampling_distribution}). As continual learning progresses we can see how the sampling from the replay buffer becomes more uneven in comparison to random sampling for the next tasks for \emph{rwd} and \emph{us}. In particular, we can see for time-steps $3$M to $4$M the detrimental effects of \emph{rwd}, since only previous experience with high rewards are sampled when learning the final task causing a lack of plasticity for \emph{rwd}. When looking at the distributions for $50:50$ sampling we can see how they are more shifted to the right in comparison to random sampling, this is to be expected since we are explicitly biasing the sampling to more recent experience to make the world model learn recent tasks quicker since $50:50$ is paired with \emph{rs}.

% \begin{figure}
%     \centering
%     \includegraphics[width=0.8\textwidth]{plots/sampling_distribution.pdf}
%     \caption{\textcolor{orange}{Distribution of episode indexes which are sampled to construct a minibatch for various sampling methods (rows) while the world-model is interacting with each task (columns).}}
%     \label{fig:sampling_distribution}
% \end{figure}

\end{appendices}

\end{document}